\documentclass{article}

\usepackage[preprint, nonatbib]{neurips_2023}
\usepackage[numbers]{natbib}

\usepackage[utf8]{inputenc} 
\usepackage[T1]{fontenc}    
\usepackage{hyperref}       
\usepackage{url}            
\usepackage{booktabs}       
\usepackage{amsfonts}       
\usepackage{nicefrac}       
\usepackage{microtype}      
\usepackage{xcolor}         
\usepackage{amsmath}
\usepackage{graphicx}
\usepackage{multirow}
\usepackage{bm}
\usepackage{wrapfig}       
\usepackage{array}

\title{Vector-based Representation is the Key: A Study on Disentanglement and Compositional Generalization}

\author{Tao Yang$^1$\thanks{Work done during internships at Microsoft Research Asia.}
, Yuwang Wang$^2$\thanks{Corresponding author}
, Cuiling Lan$^3$
, Yan Lu$^3$ 
, Nanning Zheng$^1$\\
\texttt{yt14212@stu.xjtu.edu.cn},\\
\texttt{wang-yuwang@@mail.tsinghua.edu.cn},\\
\texttt{\{culan,yanlu\}@microsoft.com},\\ \texttt{nnzheng@mail.xjtu.edu.cn}\\
$^1$Xi'an Jiaotong University, $^2$ Tsinghua University, $^3$Microsoft Research Asia
}

\begin{document}

\maketitle

\begin{abstract}
Recognizing elementary underlying concepts from observations (disentanglement) and generating novel combinations of these concepts (compositional generalization) are fundamental abilities for humans to support rapid knowledge learning and generalize to new tasks, with which the deep learning models struggle. 
Towards human-like intelligence, various works on disentangled representation learning have been proposed, and recently some studies on compositional generalization have been presented. 
However, few works study the relationship between disentanglement and compositional generalization, and the observed results are inconsistent. 
In this paper, we study several typical disentangled representation learning works in terms of both disentanglement and compositional generalization abilities, and we provide an important insight: vector-based representation (using a vector instead of a scalar to represent a concept) is the key to empower both good disentanglement and strong compositional generalization.
This insight also resonates the neuroscience research that the brain encodes information in neuron population activity rather than individual neurons.
Motivated by this observation, we further propose a method to reform the scalar-based disentanglement works ($\beta$-TCVAE and FactorVAE) to be vector-based to increase both capabilities.
We investigate the impact of the dimensions of vector-based representation and one important question: whether better disentanglement indicates higher compositional generalization.
In summary, our study demonstrates that it is possible to achieve both good concept recognition and novel concept composition, contributing an important step towards human-like intelligence. 
\end{abstract}

\section{Introduction}

Humans can effectively understand various abstract concepts from observations and efficiently generalize to a novel composition of these concepts. This remarkable ability is proposed to be an important mechanism for humans to learn knowledge and transfer it to novel contexts~\cite{cole2013rapid,frankland2020concepts,ito2022compositional}.
For example, for a human, it is easy to depict an unseen object with learned concepts such as color, shape, and texture. 
Languages generally be considered as disentangled representations for visual observations, and language can be recomposed to represent novel observations. 
Serving as a disentangled and computationally generalizable representation, languages act as powerful tools for humans to comprehend the world, learn, and create knowledge.
Similarly, it has been suggested that disentanglement~\cite{bengio2013representation} and compositional generalization~\cite{lake2017building,lake2018generalization} are fundamental missing ingredients for deep learning models to achieve human-like intelligence.

Towards this ambiguous goal, the disentangled representation learning task is proposed~\cite{bengio2013representation} to discover underlying factors/concepts behind the observations and represent each factor with explicit representations. 
Various works have been proposed for this task, and one representative branch is VAE-based~\cite{higgins2016beta, chen2018isolating, FactorVAE}. Two recent works, SAE~\cite{leebstructure} and VCT~\cite{yang2022visual}, are based on an AdaIn-like structure or Transformer to achieve disentangled, respectively. 
The VAE-based methods and SAE represent each factor with one scalar, while VCT uses a vector, i.e., a token, to represent each factor.
Those methods achieve disentanglement, but no one evaluates their compositional generalization capabilities of them. 

Compositional generalization has drawn attention recently. Montero et al.~\cite{montero2021role} evaluated compositional generalization in terms of image reconstruction or generation.
A recent work~\cite{xu2022compositional} directly evaluates the compositional generalization ability of VAE-based methods and finds VAE-based methods show bad compositional generalization ability, and better disentanglement ability does not indicate higher compositional generalization. 
However, these works only evaluate VAE-based disentangled representation learning methods, and it is necessary to consider some recent disentanglement methods to uncover the relationship between disentanglement and compositional generalization.

In this paper, we conduct a study on disentanglement and compositional generalization and reveal an important insight: vector-based representation is the key to enabling good disentanglement and strong compositional generalization. By vector-based representation, we refer to using a vector instead of a scalar to represent a factor. 
We examine the latest vector-based disentangled method VCT~\cite{yang2022visual} and find it can achieve both good disentanglement and strong compositional generalization.
Motivated by this observation, we propose a method to vectorize the representations of two popular VAE-based methods ($\beta$-TCVAE~\cite{chen2018isolating} and FactorVAE~\cite{FactorVAE}) and SAE~\cite{leebstructure}. Besides increasing the dimension of the latent vectors, we also need to reform the loss function and modify the architecture to satisfy the model's disentanglement requirement. 
The three vectorized methods demonstrate stronger compositional generalization with an average increase of 51\% with good disentanglement (some of the methods even improved) on Shapes3D compared to the scalar-based ones.
This observation is in conformity with the \emph{population coding} in neuroscience. The brain encodes information in the population activity of neurons: \emph{individual neurons count for little; it is population activity that matters}~\cite{averbeck2006neural}. 
Intuitively, scalar-based representation (i.e., single neurons) is not very informative, and vector-based representation allows for the inclusion of more information for each concept. Therefore, we experiment with increasing the number of vector dimensions and find the compositional generalization also improves. Similar observations are also studied in~\cite{xu2022compositional}, demonstrating that large bandwidth improves the compositional generalization in Emergent Language Model. We further study the relation between disentanglement and compositional generalization for the vector-based methods. We observed a positive correlation between one of the compositional generalization metrics and disentanglement.

Our main contributions can be summarized as follows:
\begin{itemize}
\item We provide an important insight that vector-based representation is one of the keys to both good disentanglement as well as strong compositional generalization.
\item We provide a vectorization method to transfer scalar-based methods into vector-based ones to unify the existing models into two categories: vector-based and scalar-based disentanglement methods.
\item We conduct experiments to reveal the relation between disentanglement and compositional generalization for vector-based methods: the compositional generalization classification metric positively correlates to the disentanglement, but the regression metric does not. 
\end{itemize}

\section{Related Works}
\subsection{Disentangled Representation Learning} The disentangled representation learning is first introduced in \citet{bengio2013representation}.  
The conventional disentangled representation is that each scalar of the representation only encodes a single independent factor, which is a scalar-based representation in this study. There are some inductive biases proposed to achieve such scalar-based disentanglement. For example, VAE-based works constraints the latent probabilistic distributions~\cite{chen2018isolating, FactorVAE, higgins2016beta, burgess2018understanding, yang2021towards, locatello2019challenging}. SAE~\cite{leebstructure} proposes to adopt a StyleGAN generator-like architecture as the structure inductive bias. However, these models are, in general, only designed for disentangled representation learning, where compositional generalization is not considered.
Few works focus on vector-based representation in disentangled representation learning~\cite{du2021unsupervised}. Although ~\cite{yang2022visual} proposes a transformer-based model to learn vector-based representation, no previous works explored its compositional generalization ability to the best of our knowledge.  
\subsection{Compositional Generalization} The compositional generalization is studied on generative models systematically in~\citet{zhao2018bias}. However, the disentanglement is not considered. The compositional generalization problem in disentangled representation learning was first studied in~\citet{esmaeili2019structured, higgins2017scan}, but there are only several specific forms of combinatorial generalizations, and the role of disentanglement on generalization is not clear. Different from ~\citet{montero2021role}, ~\citet{xu2022compositional} directly evaluates the compositional generalization and uses random train-test splits rather than manually selected splits. However, these two works are conducted on scalar-based representation and only consider the VAE-based method. Both two studies find that the disentanglement of VAE-based methods is not correlated or even inversely associated with the compositional generalization. These inspire us to ask the question: is it possible that there exists a model equipping two abilities, and how to train a model to achieve such a goal?

\section{Background: Disentanglement Models} 
\label{sec:method_scalar}
In this section, we introduce the disentanglement models used in this paper. We select two popular VAE-based disentangling models: $\beta$-TCVAE~\cite{chen2018isolating} and FactorVAE~\cite{FactorVAE}. In addition, we also consider two recently proposed disentanglement models: SAE~\cite{leebstructure}, a structure-based disentangling method, and VCT~\cite{yang2022visual}, a transformer-based method.
\begin{figure}
    \centering
    \includegraphics[width=\linewidth]{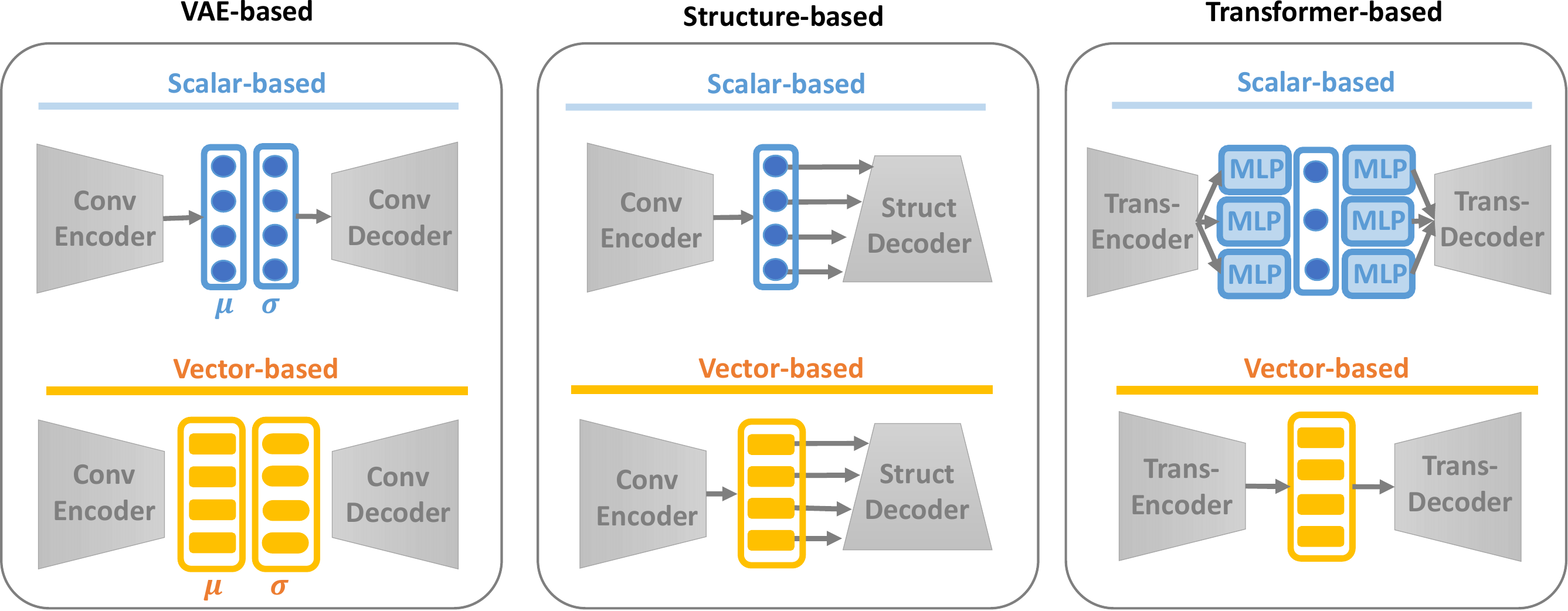}
    \caption{The unified illustration of scalar-based and vector-based disentanglement methods. For VAE-based and structure-based methods, we extend a scalar of the encoder output to a vector, where each vector represents a factor. We reformulate the loss function of the VAE-based method in Section \ref{sec:method_vector}. A series of MLPs is employed to map each vector into a scalar for the transformer-based method.}
    \vspace{-1em}
    \label{fig:method}
\end{figure}
\subsection{VAE-based Disentanglement}
The disentangled representation learning assumes that the data $x$ is generated from a set of ground truth factors $\{f_i\}_{i=1}^N$. The goal of unsupervised disentangled representation learning is to learn representation $z$ of data $x$ such that each unit $z_i$ is a function of a single factor $f_k$, where $1\leq k\leq N$. 
VAE-based methods adopt total correlation as the regularization to encourage disentanglement.

Specifically, the above two VAE-based methods decompose the total correlation from the KL regularization term of the vanilla VAE~\cite{kingma2013auto}. We thus penalize the total correlation with a hyper-parameter $\gamma$. The total loss function is:
\begin{equation}
    \mathcal{L} = \mathbb{E}_{q(z|x)p(x)}\left[p_\theta(x|z)\right] - \mathbf{KL}(q_\phi(z|x)||p(z)) - \gamma \mathbf{KL}(q_\phi(z)||\prod_i q_\phi(z_i)),
    \label{eq:vae_loss}
\end{equation}
where the last term is the total correlation, and $p(z)$ is the prior distribution $\mathcal{N}(0,I)$. 
The conditional distribution $q_\phi(z|x)$ is modeled by an encoder parameterized by $\phi$. 
The posterior $p_\theta(x|z)$ is modeled by a decoder parameterized by $\theta$. 

$\beta$-TCVAE and FactorVAE use two different ways to estimate the total correlation. $\beta$-TCVAE uses the following equation to estimate it:
\begin{equation}
\mathbf{KL}(q_\phi(z)||\prod_i q_\phi(z_i)) = \mathbb{E}_{q_\phi(z)} [\log(q_\phi(z)) -
    \log(\prod_i q_\phi(z_i))].
\label{betaTC}
\end{equation}
While the FactorVAE utilizes a discriminator $\mathcal{D}$ to approximate $q_\phi(z)$ and $\prod_iq_\phi(z_i)$. Therefore, the total correlation can be estimated as follows:
\begin{equation}
    \mathbf{KL}(q_\phi(z)||\prod_i q_\phi(z_i)) = \mathbb{E}_{q_\phi(z)} [\log(\mathcal{D}(z)) -
    \log(1-\mathcal{D}(z))],
    \label{eq:fact_loss}
\end{equation}
where the discriminator $\mathcal{D}$ is learned by adversarial training simultaneously. 
The discriminator is trained to classify between samples from $q_\phi(z)$ and $q_\phi(\bar{z})$, where $\bar{z}$ is the representation permuted along dimension $i$.

\subsection{Structure-based Disentanglement}
SAE proposes a structural decoder to learn a hierarchy of latent variables so that the encoded information can be factorized without additional regularization. As shown in Fig. \ref{fig:method}, SAE adopts an AdaIN-like structure to modulate the spatial feature to reconstruct the image, which is a similar architecture to StyleGAN~\cite{karras2019style}. However, the injection layer maps an encoded scalar rather than a vector, as did in StyleGAN.

\subsection{Transformer-based Disentanglement}
VCT use stacked cross-attention layers to induct visual information from the image without self-attention between different concepts, which prevents information leakage across units. Besides, a Concept Disentangling Loss is proposed to facilitate the mutual exclusion of different concept tokens. As shown in Fig.~\ref{fig:method}, VCT learns a vector-based disentangled representation.  

\section{Vector-based Disentangled Representation Learning}
\label{sec:method_vector}
A scalar of conventional disentangled representation encodes a single factor. We refer to this type of representation as scalar-based representation in this paper, such as VAE-based methods and SAE. Conversely, if a single factor is encoded in a vector instead, we name it the vector-based representation. In this section, we propose a \emph{vectorization} method to transfer the scalar-based method into vector-based ones, as shown in Fig.~\ref{fig:method}.

\subsection{Vectorized Representation of VAE-based Method}
Given a sample $x$, the encoder of the VAE-based method outputs scalars: mean $\mu_i$ and variance $\sigma_i$, where $i = 1,2,\dots,m$. We use $m$ to denote the number of units of the representation. 
To extend it into a vector-based one, for factor $i$, we modify the encoder to predict vectors $[\mu_{i1},...,\mu_{iD}]$ and $[\sigma_{i1},...,\sigma_{iD}]$ instead, where $D$ is the dimension of each vector. 
Since one unit encodes an individual semantic, we set the variance inside each unit $i$ the same, i.e., $\sigma_{ij} = \sigma_{ik}, j\neq k$.

The loss function (Eq.~\ref{eq:vae_loss}) should also be modified when applied to the vector-based representation. Since the first item is reconstruction loss and there is no need for modification, we focus on the last two items. 
The KL divergence can be formulated as follows:
\begin{equation}
    \mathbf{KL}(q_\phi(z|x)||p(z)) = -0.5(-\frac{1}{D}\sum_{ij}\mu_{ij}^2 - \sum_{i}(\sigma_{i} - \log \sigma_{i}) - m) \cdot D.
    \label{eq:KL_specific}
\end{equation}
Please refer to Appendix A for detailed derivation.
To make it comparable to the vanilla VAE, we ignore the multiplier $D$ when calculating the loss. We also modify the total correlation of $\beta$-TCVAE. Specifically, since the total correlation of vector-based $\beta$-TCVAE with a shared variance inside each unit is intractable, we average the total correlation along the dimension of the representation vector to approximate the total correlation of vector-based $\beta$-TCVAE:
\begin{equation}
\mathbf{KL}(q_\phi(z)||\prod_i q_\phi(z_i)) =  \frac{1}{D} \sum_j\mathbf{KL}(q_\phi(z_j)||\prod_i q_\phi(z_{ij})).
\label{eq:betaTC_approx}
\end{equation}


Note that we use average but not sum operation to make the value comparable to the original $\beta$-TCVAE. Take Eq. \ref{eq:betaTC_approx} and Eq. \ref{eq:KL_specific} together. We can obtain the loss function of $\beta$-TCVAE. Different from $\beta$-TCVAE, we need to modify the discriminator $\mathcal{D}$ of FactorVAE to estimate the total correlation of a set of joint distributions. We extend the discriminator $\mathcal{D}$ to take $z_{ij}$ as input. The permutation is only performed on dimension $i$ when we train the discriminator. Together with Eq. \ref{eq:fact_loss} and Eq. \ref{eq:KL_specific}, we can compute the loss of vectorized FactorVAE.

\subsection{Vecterized Representation of Structure-based Method}
As mentioned above, to extend the structure-based method, SAE, to be a vector-based method, we only need to replace the encoded scalar with an encoded vector of $D$ dimension, as shown in Fig.~\ref{fig:method}. Therefore, the scale and shift coefficients are predicted by a vector instead, of which the structure is the same as the generator of StyleGAN~\cite{karras2019style}. Since SAE is a regularization-free method, there is no need for modification of the loss function. 

\subsection{Vecterized Representation of Transformer-based Method}
Since VCT, a transformer-based method, is already a vector-based method, in order to unify these models and analyze two types of methods, we modify VCT into a scalar-based one. 
Specifically, as shown in Fig.~\ref{fig:method}, we use different MLP layers to map different vectors into scalars to maintain the independence of learned vectors. 
As this modification dose not affect the loss function, we keep the loss function of VCT.

\section{Experiment design}

\subsection{Dataset}
In this study, we are especially interested in exploring the disentanglement and compositional generalization ability. There are some datasets commonly used in disentangled representation learning~\cite{yang2022visual,leebstructure}. In addition, these datasets recently are also used in compositional generalization literature~\cite{xu2022compositional}. Therefore, we follow~\citet{xu2022compositional} to use two public datasets: Shapes3D~\cite{FactorVAE} and MPI3D-Real (MPI3D in short)~\cite{mpi-toy}. The Shapes3D dataset is an image dataset that is generated by six different factors: floor color, background color, object color, object size, object shape, and azimuth. MPI3D contains images synthesized with a robot arm in a controlled environment, which has seven different factors.

\textbf{Data Splits} The goal of compositional generalization is that the novel combination of seen concepts can be recognized in the downstream task. Therefore, we follow~\cite{xu2022compositional} to split the dataset into two parts: train and test (1:9). The training set is smaller than ~\citet{montero2021role, schott2021visual}. 
\subsection{Hyper-parameters} The hyper-parameters used in models used in this study are adopted by following the prior works. For $\beta$-TCVAE and FactorVAE, we adopt the implementation in \texttt{disentanglement\_lib}~\cite{locatello2019challenging}. In addition, we set the regularization strength $\gamma$ to 10, which is both used in~\citet{FactorVAE} and \citet{chen2018isolating}. For SAE, we follow~\citet{leebstructure} to use SAE-12 as the model architecture. Note that there is no regularization term in the loss function of SAE. For VCT, we follow~\citet{yang2022visual} to set the regularization coefficient of VCT as 1. Please note that we only train the VQ-VAE of VCT on the training set for a fair comparison. The training batch size is 32, and the optimizer is Adam, with a learning rate of $10^{-4}$. For more details, please refer to Appendix B.

\subsection{Evaluation Metrics}
\textbf{Disentanglement Evaluation} An ideal disentangled representation should be disentangled both on the training and testing sets. In this study, we follow~\citet{xu2022compositional} to focus on the performance of the testing set, which indicates how the model is able to disentangle the unseen combination of factors. We also follow them to randomly split the dataset by using $3$ random seeds. Conventionally, the disentanglement is often influenced by randomness. Therefore, we follow~\cite{locatello2019challenging} to conduct our experiments with $5$ random seeds for each splitting random seed. We have $15=5\times 3$ runs for each method on each dataset. Following \citet{yang2022visual}, four popular metrics are used in our experiments: the FactorVAE score~\cite{FactorVAE}, the DCI~\cite{eastwood2018framework}, the $\beta$-VAE score~\cite{higgins2016beta}, and MIG~\cite{chen2018isolating}. We follow \citet{du2021unsupervised,yang2022visual} to evaluate the performance of vector-based representation with these metrics.

\textbf{Compositional Generalization Evaluation} ~\citet{xu2022compositional} evaluates the compositional generalization by testing how easily a simple model can predict the ground truth of factors of novel combinations. We follow \citet{xu2022compositional} to train a simple classifier and aggressor on top of the learned representation. Specifically, the ridge regression model is used for regression and logistic regression for classification. Therefore, the metrics are $R^2$ score (R2) and classification accuracy (ACC). We also follow \citet{xu2022compositional} to use $N_{label}=500$ labeled data to train the simple classifier or aggressor.

\section{Key Study and Results}
\label{gen_inst}

\begin{table*}[t]
\vspace{-1em}
\caption{Comparisons of disentanglement and compositional generalization between the scalar-based and vector-based methods (mean $\pm$ std, higher is better). 
Vector-based methods achieve better performance with a large margin than scalar-based methods in terms of compositional generalization. For vec-VCT*, $D=256$ is the same as ~\citet{yang2022visual}. More results are in Appendix C.}
\begin{center}
\resizebox{\textwidth}{!}{
\begin{tabular}{ccccc|cccc}
\toprule
\multirow{2}*{\textbf{Method}} & \multicolumn{4}{c}{Shapes3D} & \multicolumn{4}{c}{MPI3D} \\
\cmidrule(lr){2-9}
& FactorVAE & DCI & R2 & ACC & FactorVAE & DCI & R2 & ACC \\
\midrule
\multicolumn{9}{c}{\textit{Scalar-based:}} \\
\midrule
FactorVAE & $0.83 \pm 0.06$ & $0.44 \pm 0.12$ & $0.46 \pm 0.18$ & $0.39 \pm 0.10$ & $0.31 \pm 0.04$ & $0.21 \pm  0.01$ & $0.30 \pm 0.02$ & $0.39 \pm 0.02$ \\
$\beta$-TCVAE & $0.83 \pm 0.10$ & $0.65 \pm 0.16$ & $0.45 \pm 0.15$ & $0.47 \pm 0.18$ & $0.44 \pm 0.05$ & $0.27 \pm 0.01$ & $0.32 \pm 0.03$ & $0.45 \pm 0.03$ \\
SAE & $0.98 \pm 0.04$ & $0.87 \pm 0.12$ & $0.72 \pm 0.05$ & $0.90 \pm 0.17$ & $0.71 \pm 0.04$ & $0.47 \pm 0.05$ & $0.55 \pm 0.07$ & $0.77 \pm 0.02$ \\
VCT & $0.95 \pm 0.05$ & $0.86 \pm 0.02$ & $0.56 \pm 0.24$ & $0.58 \pm 0.15$ & $\bm{0.72 \pm 0.04}$ & $0.47 \pm 0.03$ & $0.39 \pm 0.13$ & $0.69 \pm 0.09$ \\
\midrule
\multicolumn{9}{c}{\textit{Vector-based:}} \\
\midrule
vec-FactorVAE & $0.93 \pm 0.06$ & $0.55 \pm 0.11$ & $0.88 \pm 0.05$ & $0.96 \pm 0.02$ & $0.38 \pm 0.06$ & $0.16 \pm 0.05$ & $0.53 \pm 0.02$ & $0.71 \pm 0.01$ \\
vec-$\beta$-TCVAE & $0.82 \pm 0.08$ & $0.31 \pm 0.08$ & $0.87 \pm 0.05$ & $\bm{0.98 \pm 0.01}$ & $0.42 \pm 0.06$ & $0.11 \pm 0.03$ & $0.67 \pm 0.02$ & $0.78 \pm 0.01$ \\
vec-SAE & $0.89 \pm 0.08$ & $0.63 \pm 0.06$ & $0.95 \pm 0.01$ & $0.98 \pm 0.01$ & $0.62 \pm 0.08$ & $0.33 \pm 0.09$ & $\bm{0.87 \pm 0.03}$ & $\bm{0.88 \pm 0.01}$ \\
vec-VCT & $\bm{0.98 \pm 0.04}$ & $0.85 \pm 0.06$ & $0.91 \pm 0.10$ & $0.80 \pm 0.09$ & $0.70 \pm 0.06$ & $\bm{0.48 \pm 0.04}$ & $0.70 \pm 0.07$ & $0.77 \pm 0.02$ \\
vec-VCT* & $0.97 \pm 0.04$ & $\bm{0.89 \pm 0.02}$ & $\bm{0.99 \pm 0.02}$ & $0.90 \pm 0.03$ & $0.66 \pm 0.03$ & $0.45 \pm 0.06$ & $0.85 \pm 0.07$ & $0.78 \pm 0.02$ \\

\bottomrule
\end{tabular}}
\end{center}
\vspace{-1.5em}
\label{tbl:dis_qnti}
\end{table*}

\subsection{Vector-based Representation Can Posses Both Disentangling and Compositional Generalization}
In this section, we compare the disentanglement and generalization of different models with scalar-based and vector-based representation. We train the models introduced in Sec. \ref{sec:method_scalar} and \ref{sec:method_vector} on Shapes3D and MPI3D. Tab. \ref{tbl:dis_qnti} shows the disentanglement and generalization performance. We highlight the following observations:

\textbf{Vector-based representation} Comparing the vector-based methods to scalar-based ones with the same inductive bias, the generalization performance of vector-based methods is always better than scalar-based ones, no matter which kind of method is used. Since the inductive bias still maintains by the vector-based method, there is only some performance drop on disentanglement. Even in some cases, e.g., FactorVAE on Shapes3D, there is a performance gain for vector-based FactorVAE. Since we use an approximated total correlation, the performance of vec-$\beta$-TCVAE significantly drops. 

\textbf{Different disentangling method} One can observe that there is no trade-off between the disentanglement and generalization for all the methods. vec-VCT achieves SOTA performance on both disentanglement and compositional generalization. As for vector-based methods, we observe that better disentangling methods perform better on generalization. 

\textbf{Implications} Our conclusion is different from~\citet{xu2022compositional}, which concludes that better-disentangled representation  produces representations with worse generalization. This study is conducted on the scalar-based disentangling method. 
While for vector-based methods, better disentanglement will not lead to worse generalization performance. 
Compared to the scalar-based methods, vector-based methods increase the bandwidth of the bottleneck and enhance the generalization performance, which is consistent with the conclusion of the EL model in ~\citet{xu2022compositional}. 
In order to further analyze the influence of the bottleneck bandwidth on disentanglement and generalization ability, we also train models with different vector sizes.

\subsection{Large Vector Size Results in Better Performance for Both Abilities}
\begin{figure}[t]
\centering
\begin{tabular}{c@{\hspace{0.1em}}c@{\hspace{0.1em}}c@{\hspace{0.1em}}c}
\multicolumn{4}{c}{\includegraphics[width=0.64\linewidth]{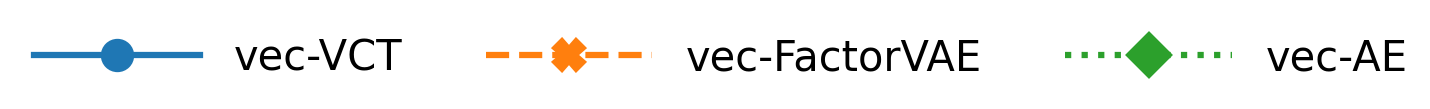}}\\
\includegraphics[width=0.24\linewidth]{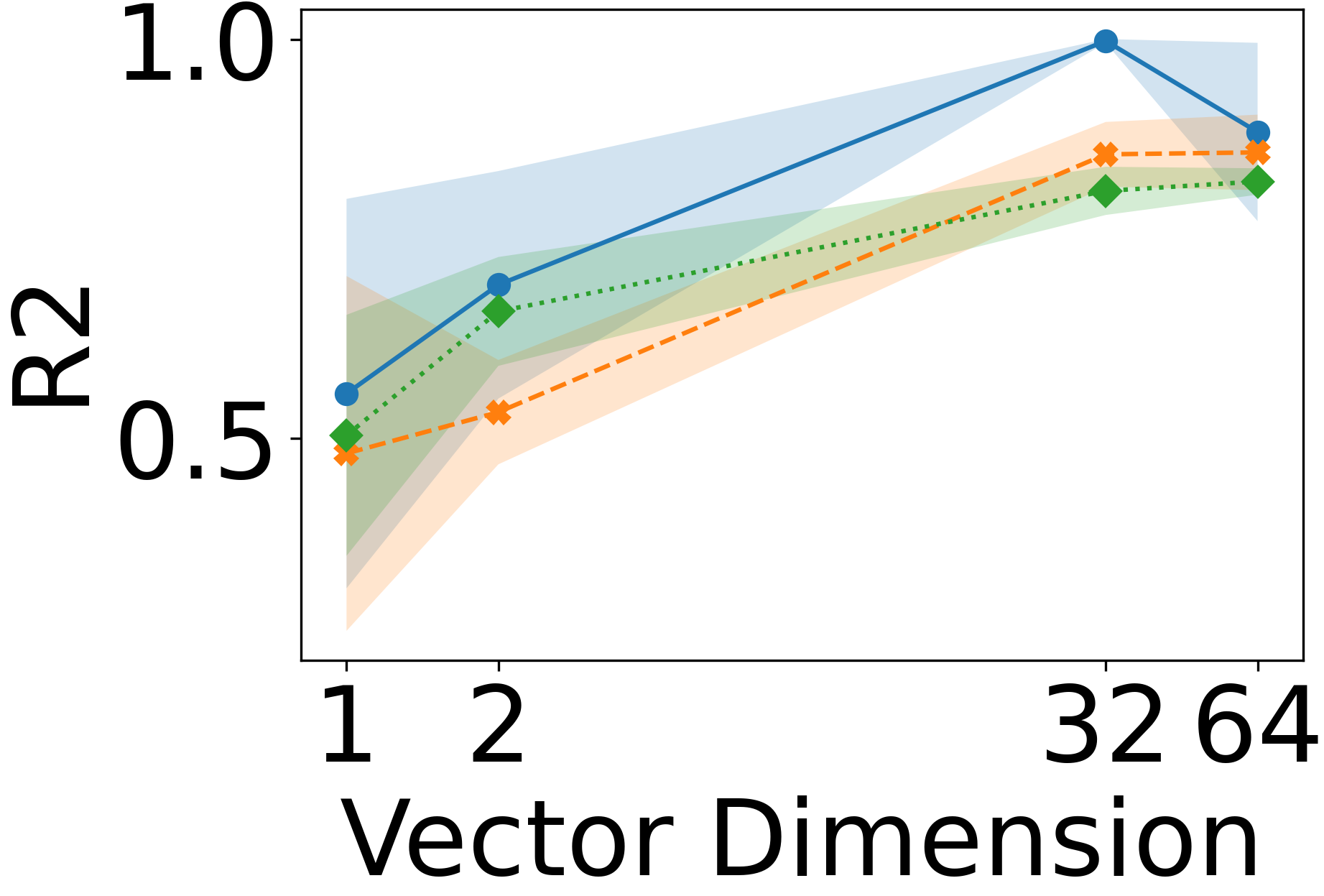} & \includegraphics[width=0.24\linewidth]{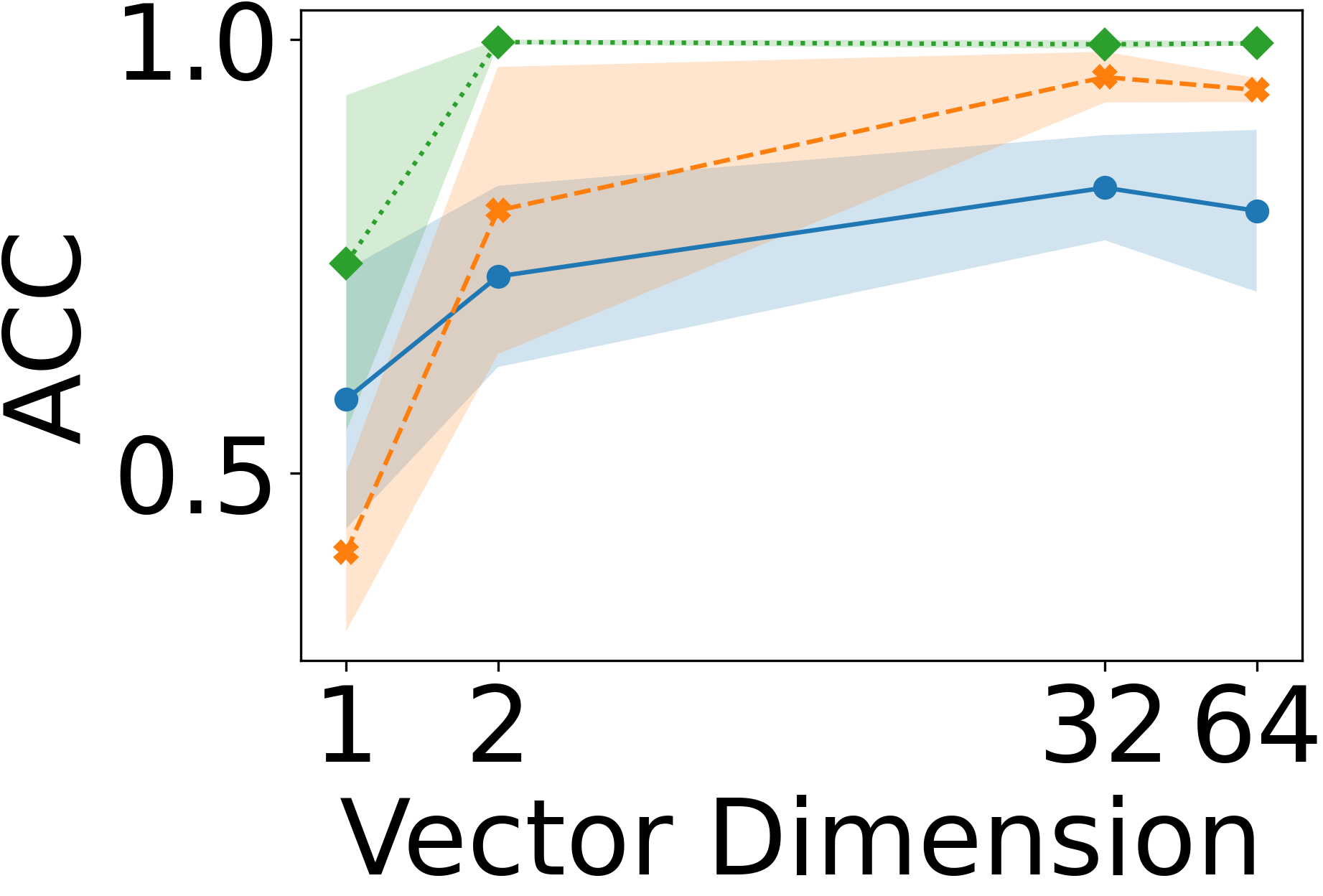} & 
\includegraphics[width=0.25\linewidth]{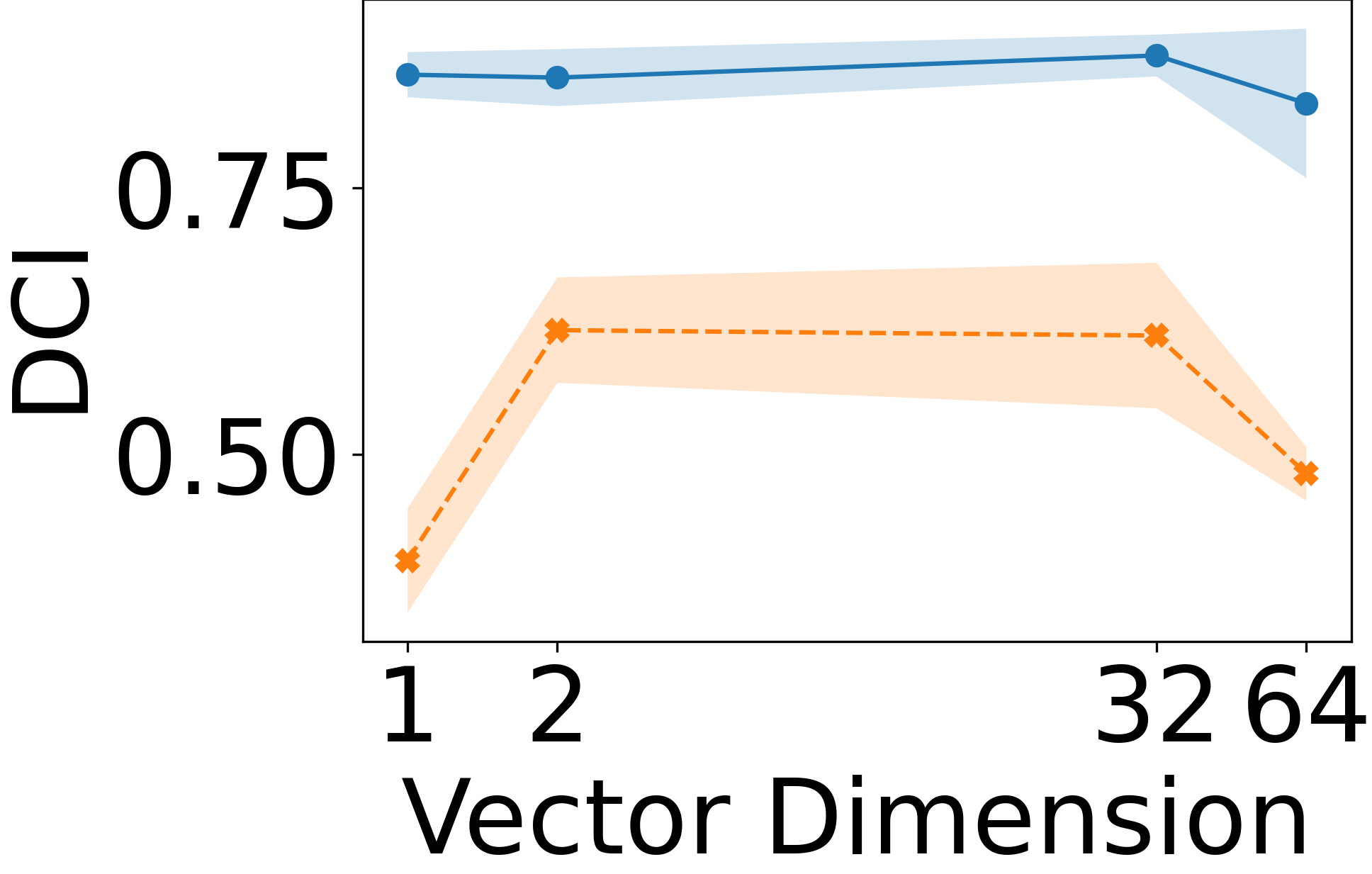} & \includegraphics[width=0.24\linewidth]{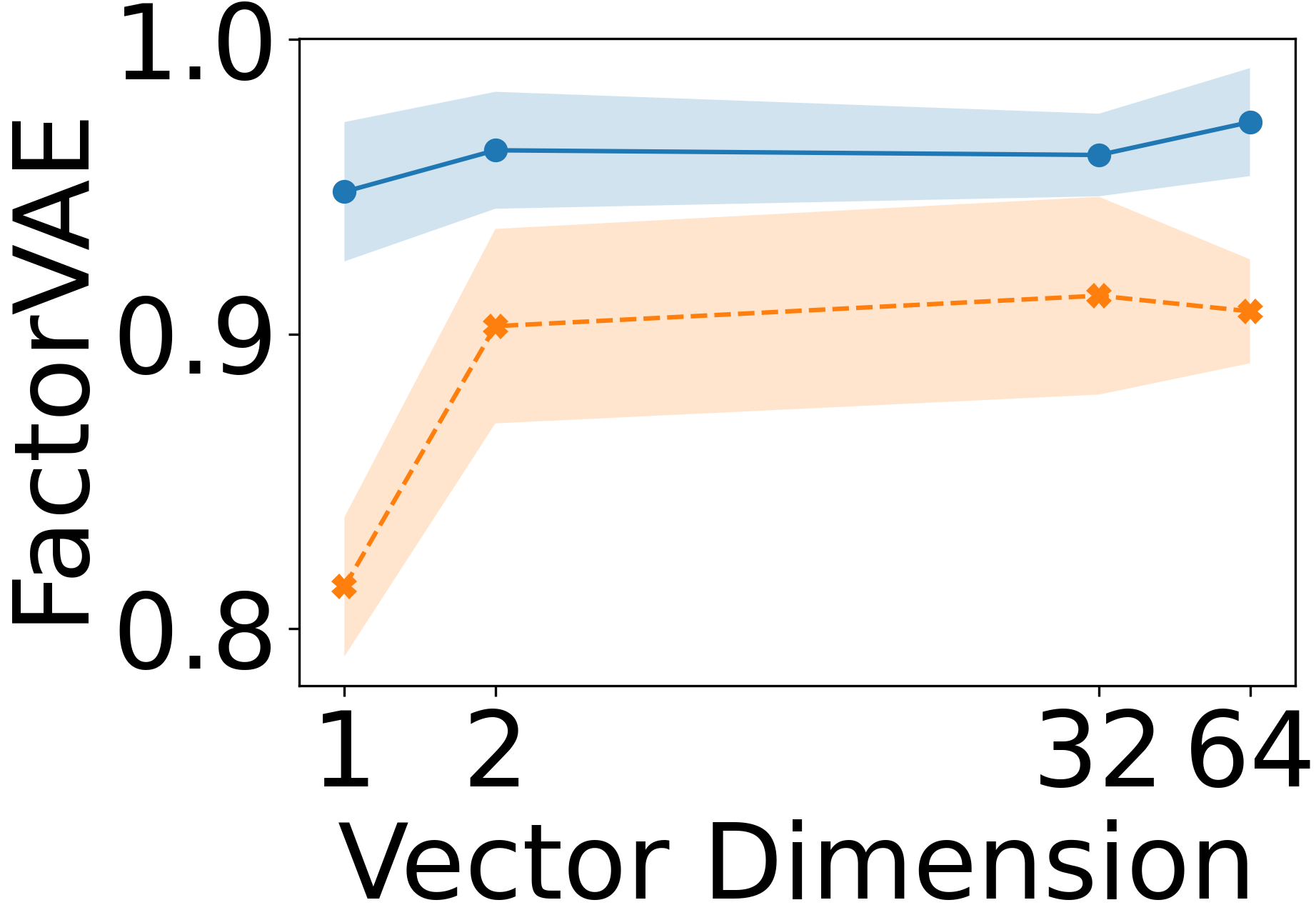}
\end{tabular}
\caption{Generalization and disentanglement performance vs vector size on Shapes3D. vec-VCT, vec-FactorVAE, and vec-AE (with regularization strength $\gamma=10$) are evaluated. The generalization metrics (R2 and ACC) are positively correlated to the vector size (vector dimension).}
\label{fig:vec_size}
\vspace{-1em}
\end{figure}
In this section, we are interested in vector-based representation with different vector sizes. Intuitively, a large vector size means large bandwidth of the bottleneck. 
We use the vec-FactorVAE and vec-VCT to conduct this experiment on Shapes3D. We also train vanilla AE for comparison. 
We evaluate the performance of the above models with different vector sizes $\{1,2,32,64\}$. We follow the prior works to set the number of units in each model to $10$. We thus set the number of latent dimensions of AE to $\{10,20,320,640\}$ so that the results of AE can be fairly compared. 

The results are shown in Fig. \ref{fig:vec_size}. As the vector size increases, the generalization performance gains consistently for all of the models. However, if the vector size is large enough, the performance also slightly drops (larger than 32). 
On the other hand, disentanglement inductive bias hurts classification but is beneficial to regression performance. 
For disentanglement performance, the increase in vector size also leads to better disentangling performance. 

\textbf{Implications} The increase of bandwidth of the bottleneck enhances the generalization performance. 
However, the increase in vector size also introduces the complexity of the latent space, and the performance drops after the vector size is larger than $32$. 
Although the models behave similarly in disentangling and generalization in this experiment, the relation between the two abilities is still unclear. For example, vec-FactorVAE and vec-VCT behave differently between $R^2$ and ACC compared to AE. We further discuss the relationship between the two abilities.

\subsection{Relation Between Disentanglement and Compositional Generalization}
We further discuss the relationship between these two abilities for vector-based representation. Since the results presented above are across different models, we study the performance with a certain type of model in this section. 
Considering that the regularization strength is correlated to the disentanglement performance, we train vec-FactorVAE and vec-$\beta$-TCVAE with different regularization strengths. Specifically, we follow ~\citet{FactorVAE} and ~\citet{chen2018isolating} to set the regularization strength to $\{5,10,20\}$ for both models. Besides, to exclude the influence of the regularization strength, we also consider the performance of models trained with different random seeds. We evaluate the trained vec-$\beta$-TCVAE, vec-FactorVAE, and vec-VCT and calculate the correlation between disentanglement and compositional generalization performance.

\begin{figure}[t]
\centering
\begin{tabular}{c@{\hspace{0.1em}}c@{\hspace{0.1em}}c@{\hspace{0.1em}}c}
\multicolumn{4}{c}{\includegraphics[width=0.54\linewidth]{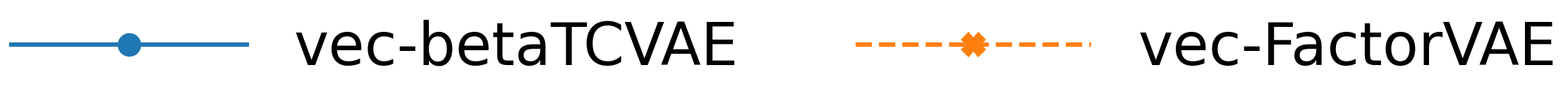}}\\
\includegraphics[width=0.24\linewidth]{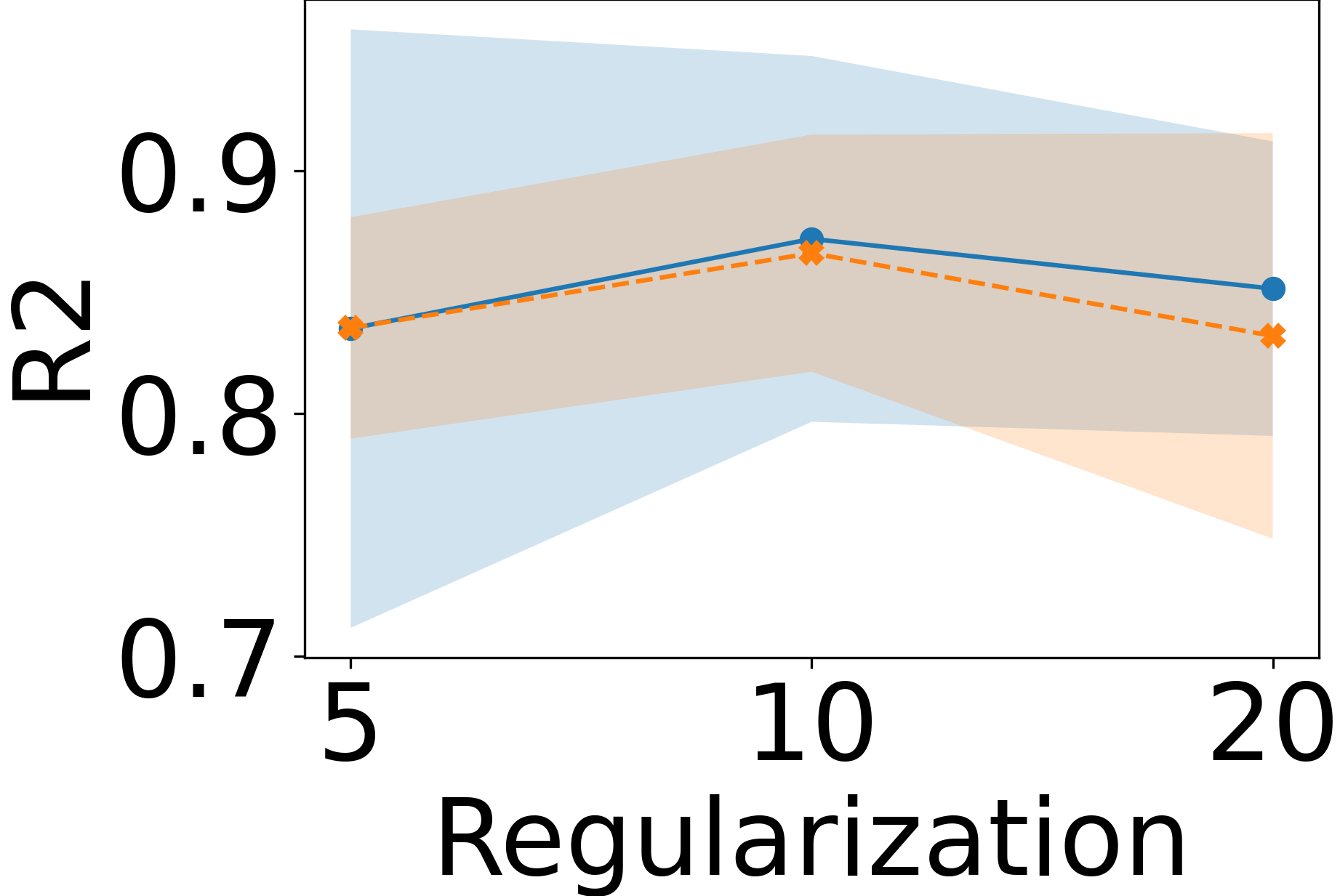} & \includegraphics[width=0.24\linewidth]{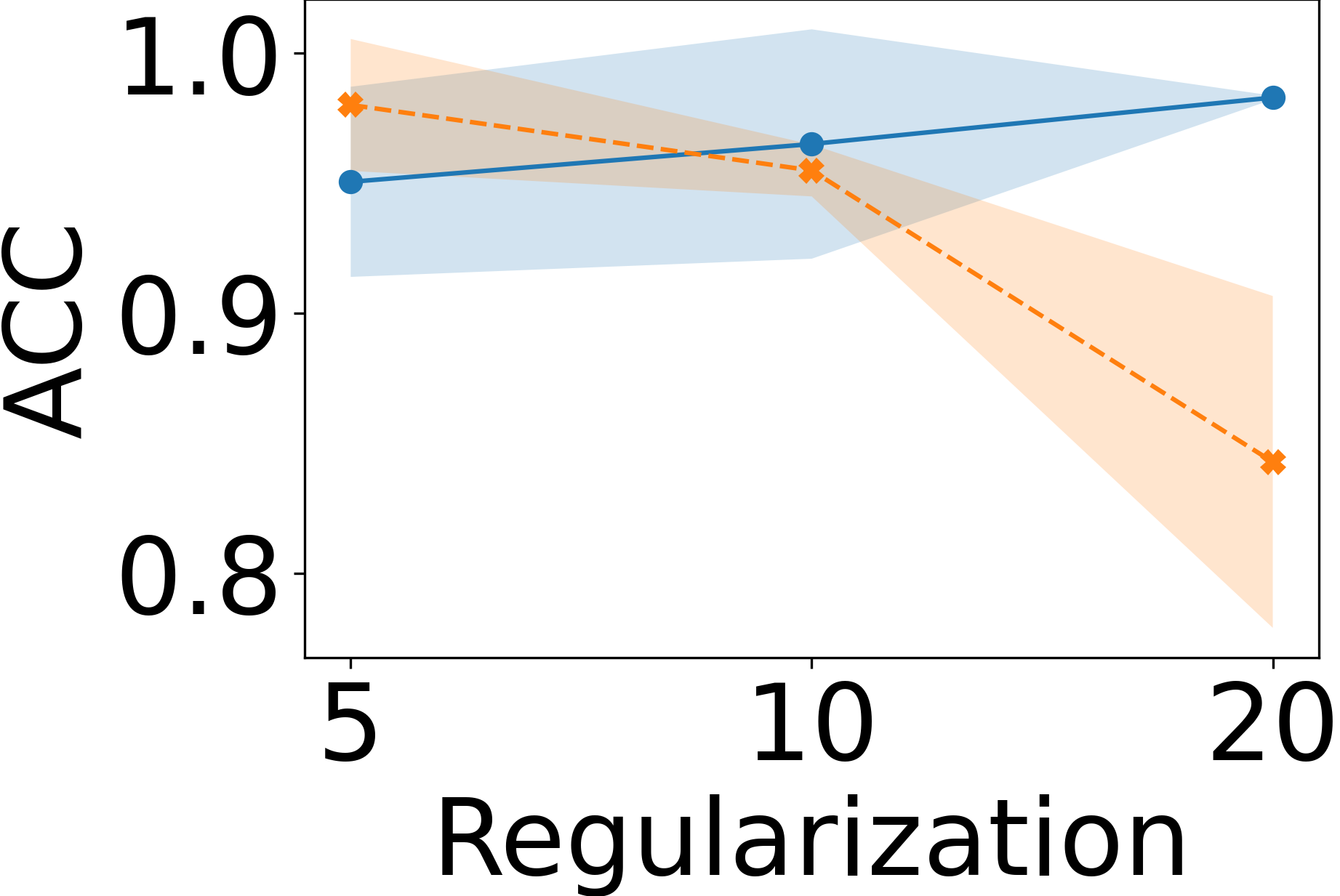} & 
\includegraphics[width=0.24\linewidth]{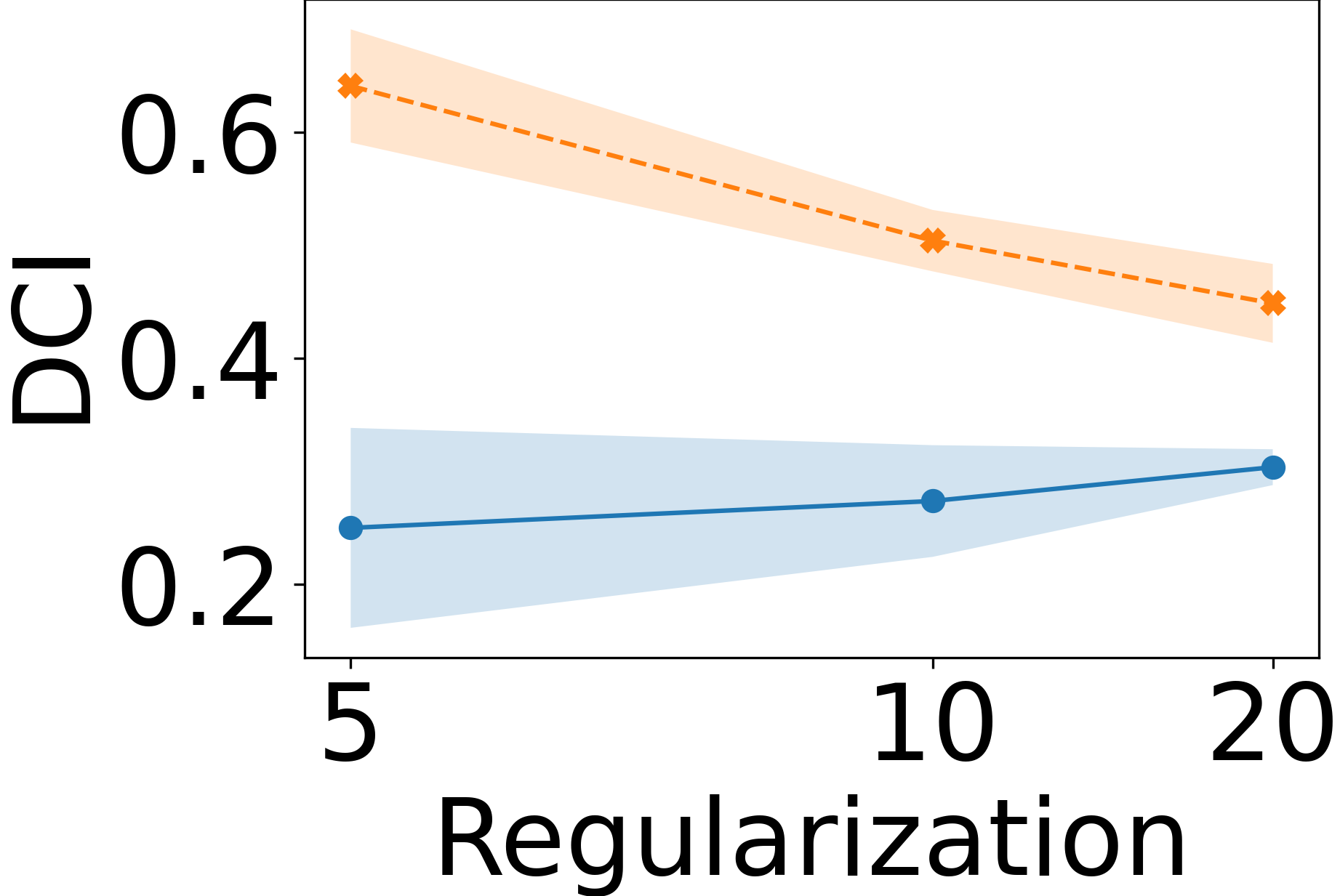} & \includegraphics[width=0.24\linewidth]{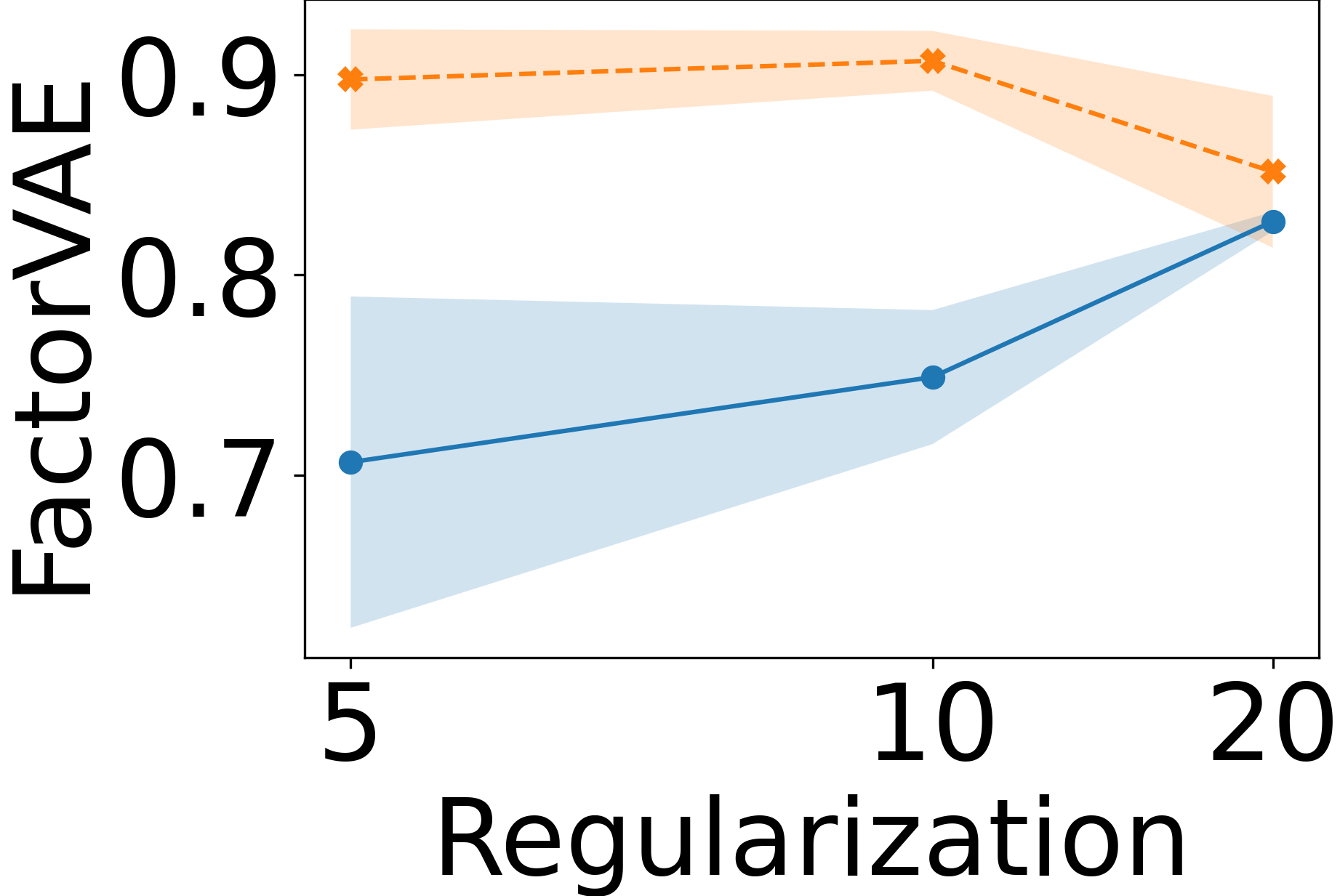} 
\end{tabular}
\caption{Generalization and disentanglement performance vs Regularization strength on Shapes3D. Two vector-based methods (with vector size $D=64$) are evaluated. vec-betaTCVAE and vec-FactorVAE use varying regularization strength $\gamma \in \{5,10,20\}$.}
\label{fig:vec_lam}
\vspace{-1em}
\end{figure}

\begin{figure}[t]
\centering
\begin{tabular}
{m{3cm}<{\centering}m{3cm}<{\centering}m{3cm}<{\centering}m{3.5cm}}
\includegraphics[width=\linewidth]{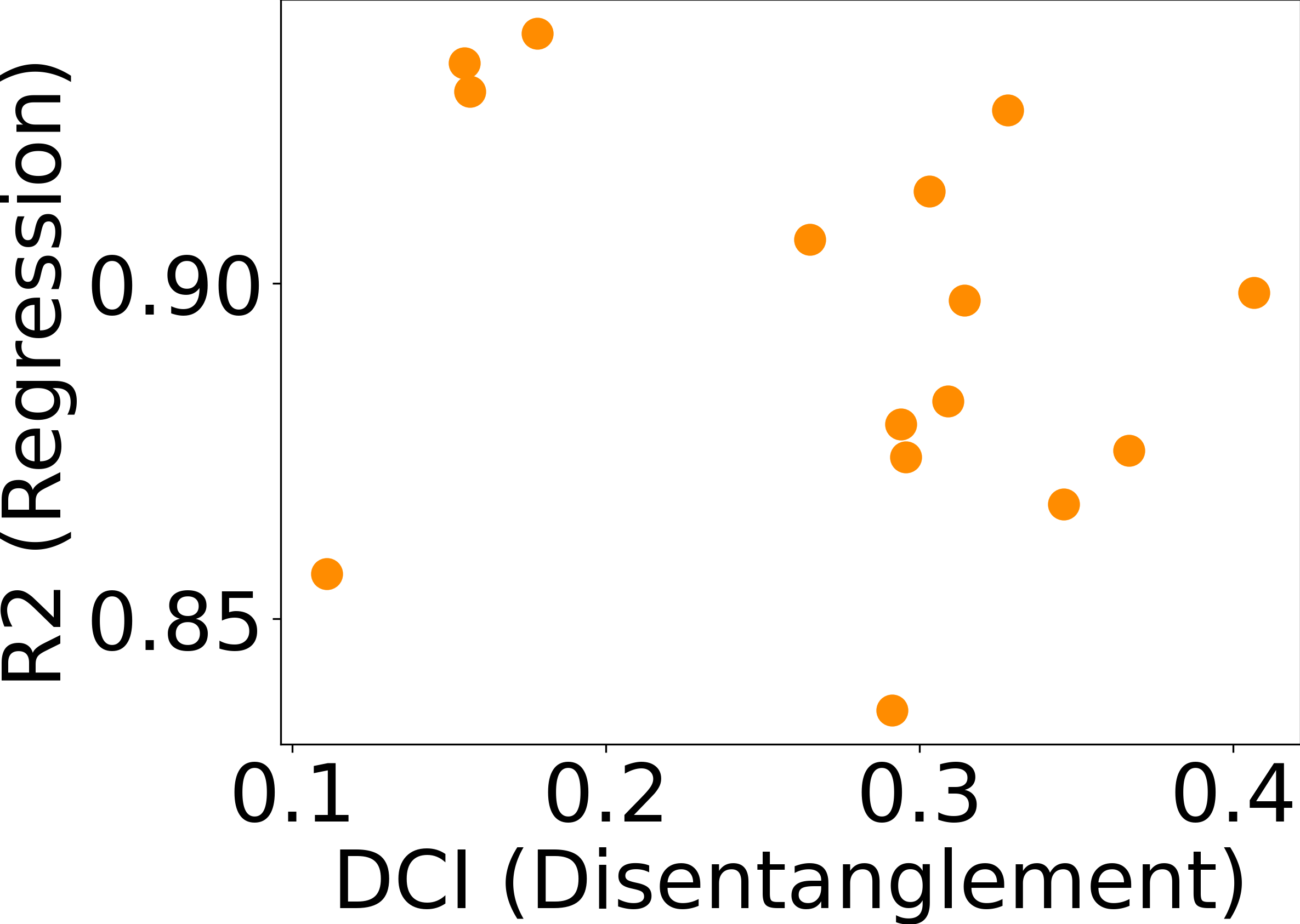} & \includegraphics[width=\linewidth]{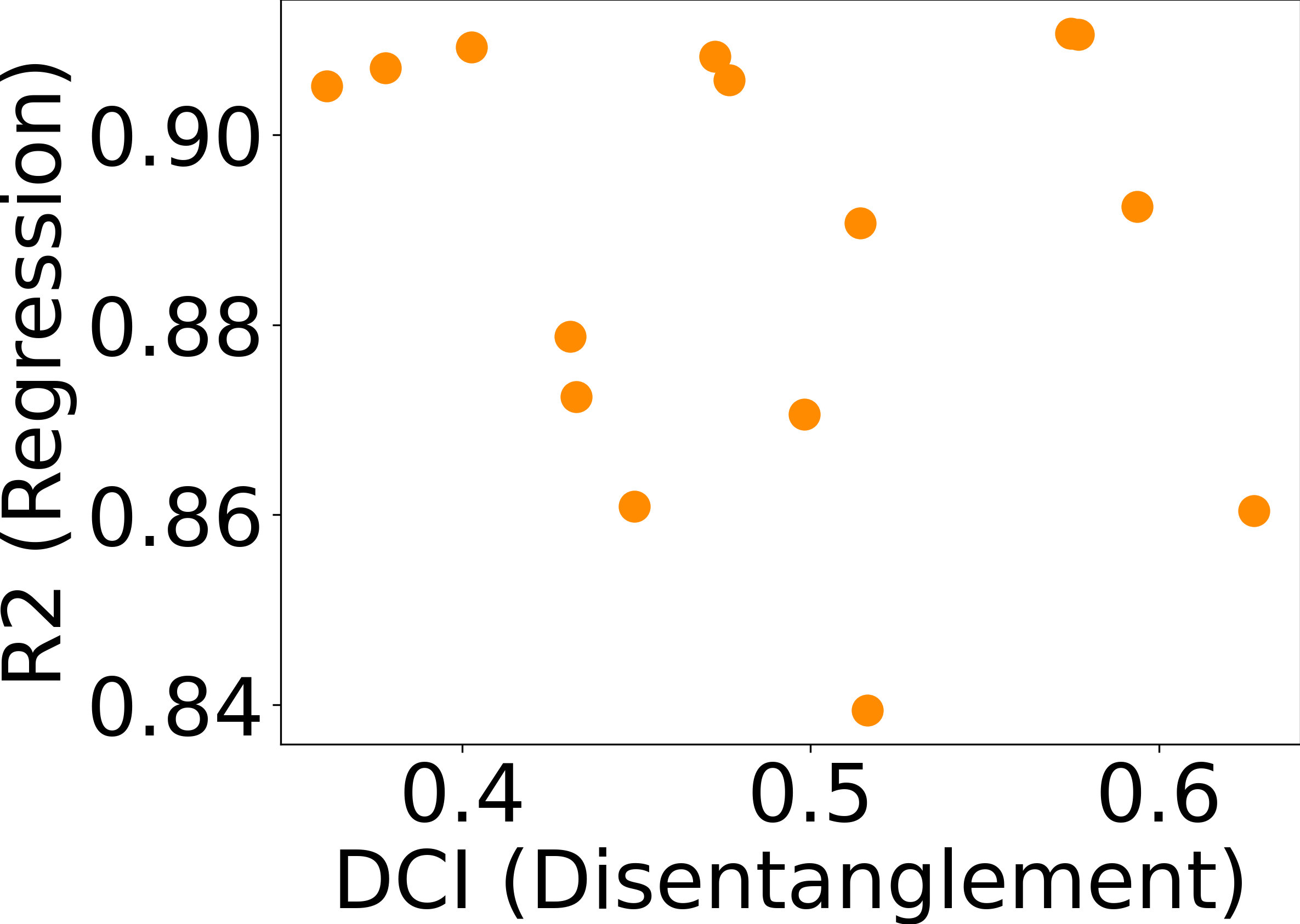} & 
\includegraphics[width=\linewidth]{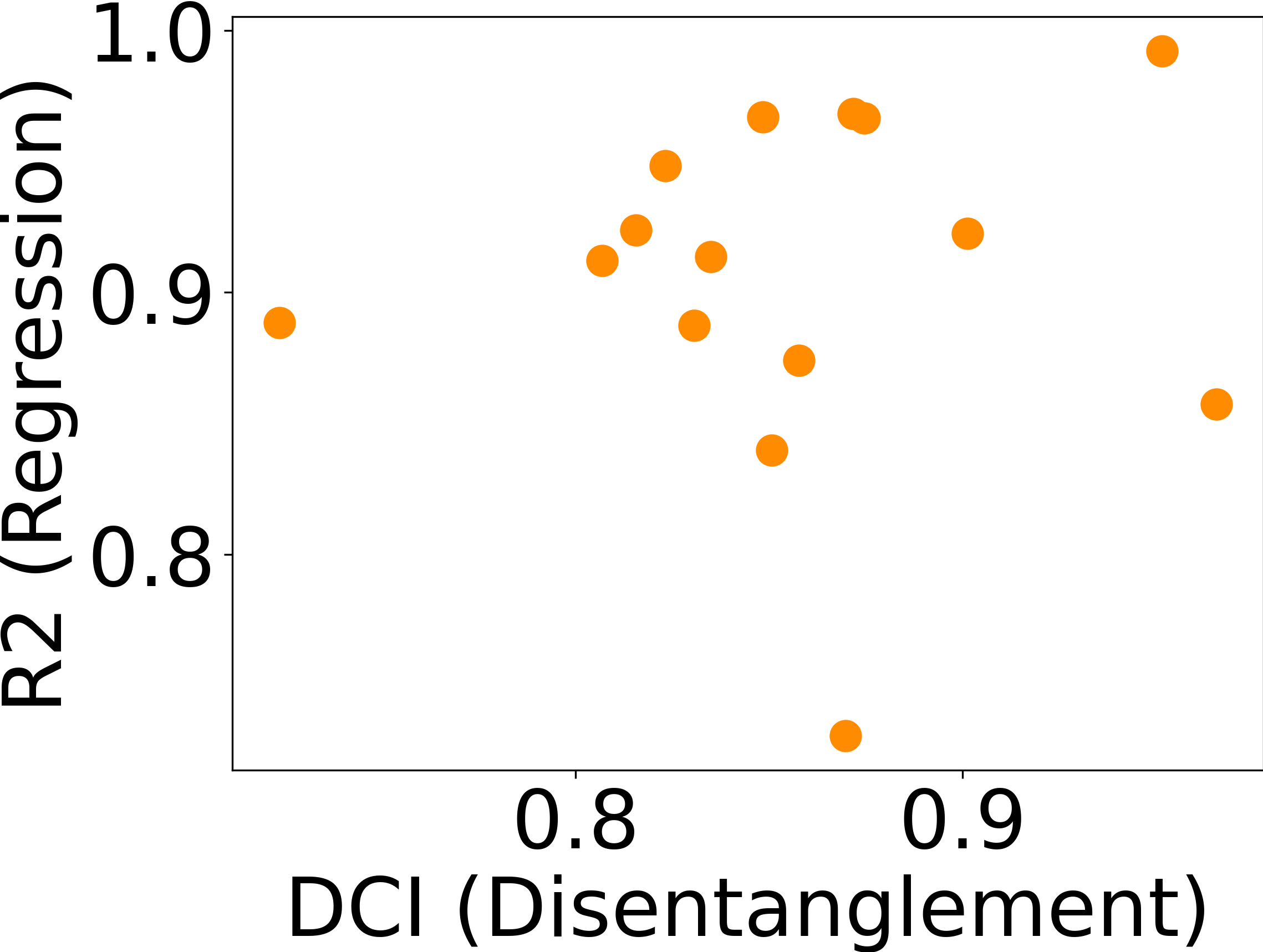} & \multirow{2}*{\resizebox{0.95\linewidth}{!}{ 
\begin{tabular}{lc}
\toprule
\multicolumn{1}{c}{Method} & Correlation\\
\midrule
\multicolumn{2}{c}{\textit{Regression:}} \\
\midrule
vec-$\beta$-TCVAE             & $-0.266$   \\
vec-FactorVAE               & $-0.194$ \\ 
vec-VCT     & $0.063$  \\
\midrule
\multicolumn{2}{c}{\textit{Classification:}} \\
\midrule
vec-$\beta$-TCVAE             & $0.824$   \\
vec-FactorVAE               & $0.638$ \\ 
vec-VCT    & $0.675$  \\
\bottomrule
\end{tabular}}}\\
\includegraphics[width=\linewidth]{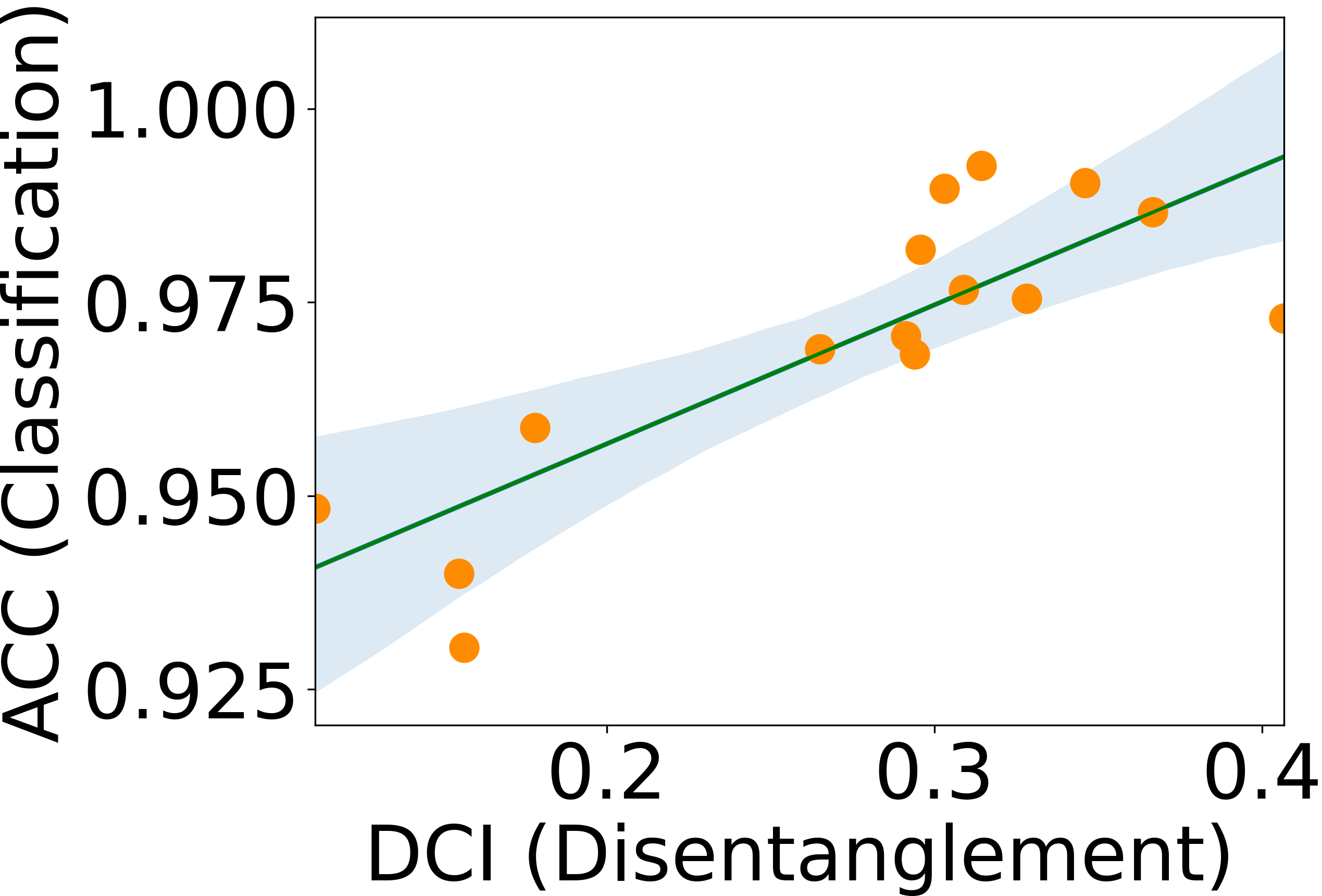} & \includegraphics[width=\linewidth]{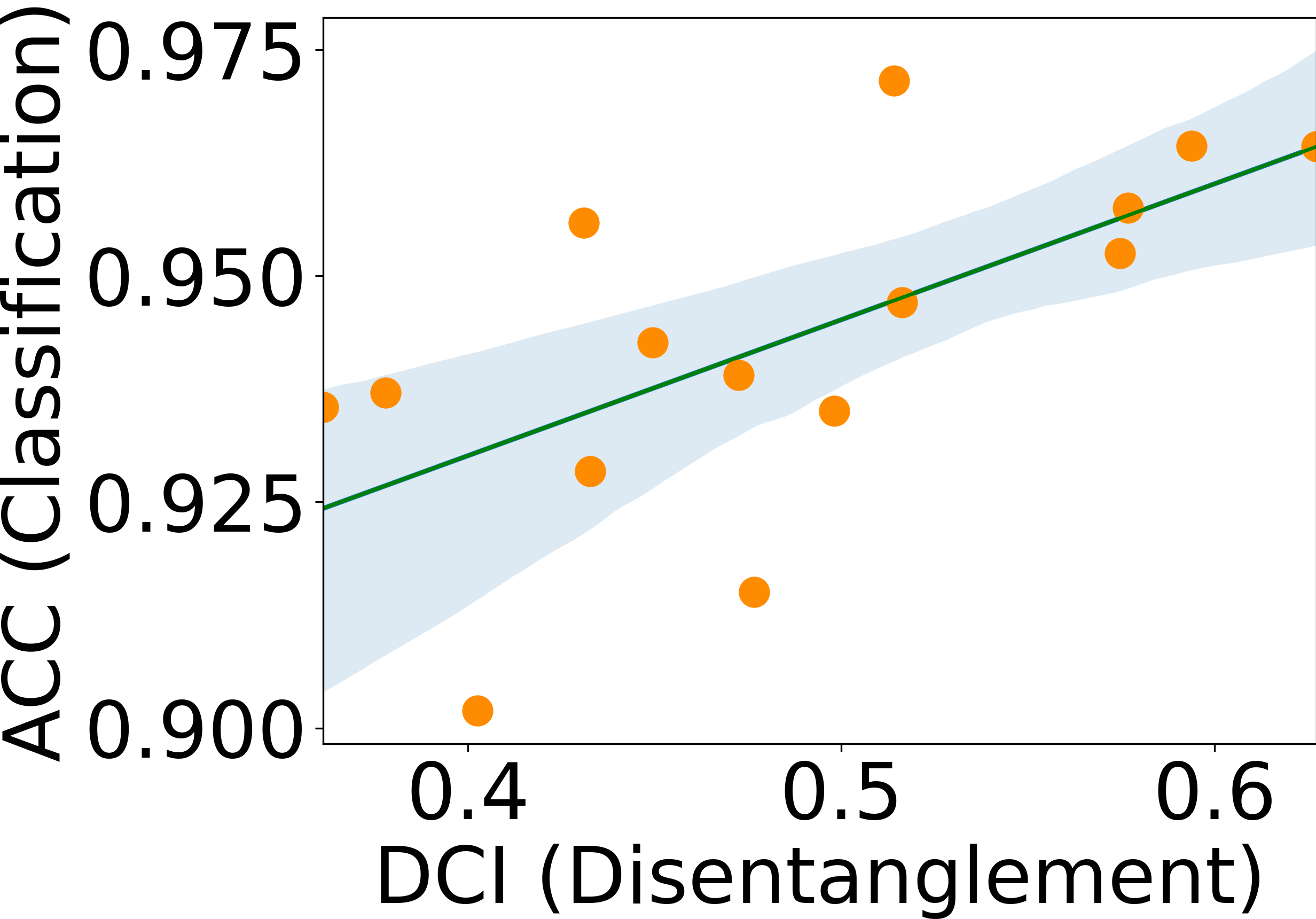} & 
\includegraphics[width=\linewidth]{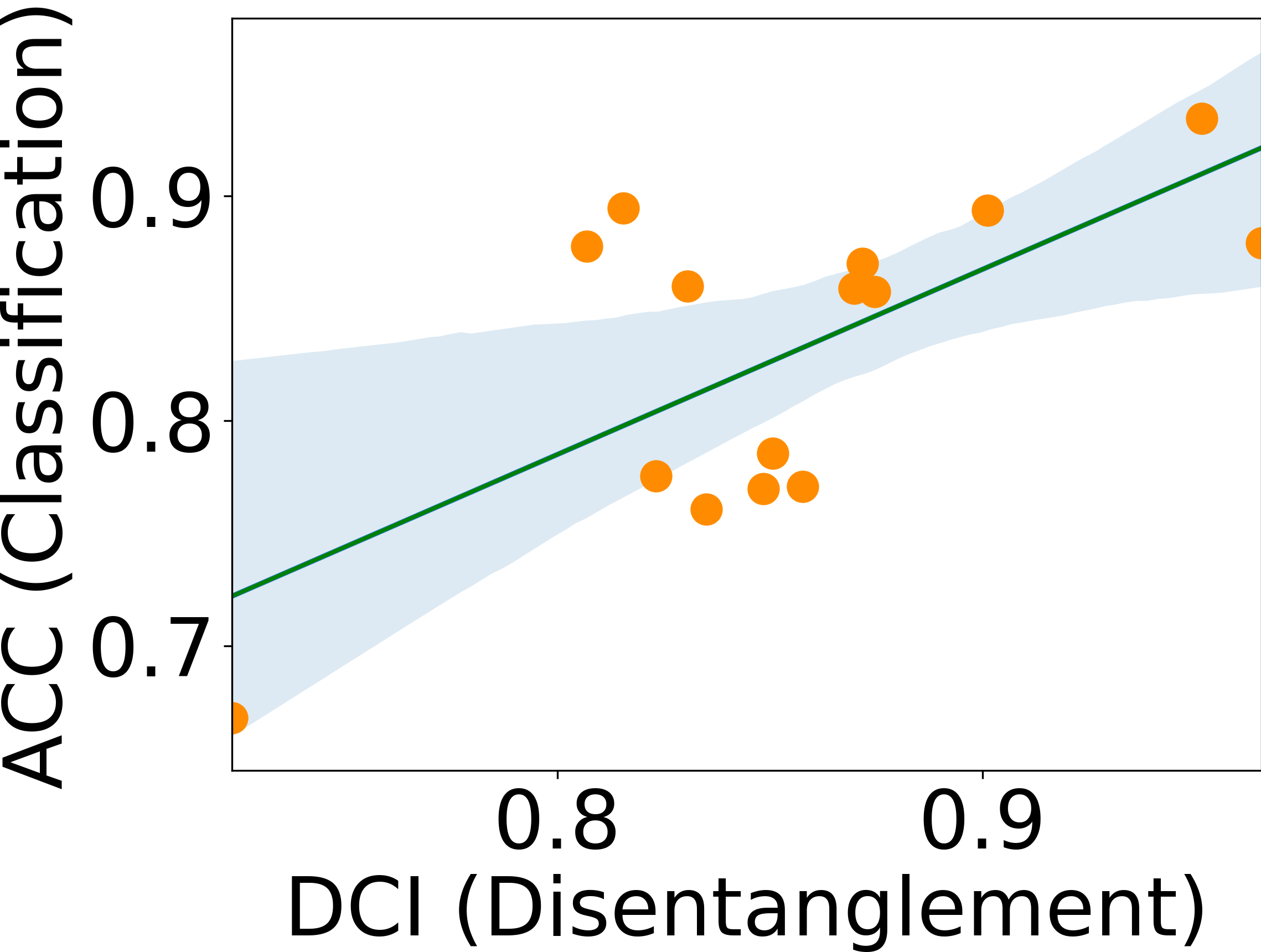} \\
vec-$\beta$-TCVAE & vec-FactorVAE  & vec-VCT & \quad Pearson Correlation \\
\end{tabular}
\caption{Generalization performance vs disentanglement performance on the Shapes3D with $D=64$ and $\gamma=10$. The Pearson correlation coefficient is calculated in the table. Orange data points represent instances of the same vector-based method trained using different random seeds.}
\label{fig:vec_random}
\end{figure}

The experiment results are shown in Fig. \ref{fig:vec_lam}. We can observe that as the regularization strength increase, the disentanglement drops for vec-FactorVAE but improves for vec-$\beta$-TCVAE. Although the generalization performance behaves similarly, there is only little influence on the regression metric. In addition, except for the changes in performance, the performance variance is reduced by the increase of the regularization strength for $\beta$-TCVAE. The performance of models trained with the same regularization strength is demonstrated in Fig. \ref{fig:vec_random}. The performance of classification and disentanglement show a positive correlation, while no significant correlation is observed for regression. 
To further confirm the relationship, we calculated the Pearson correlation coefficient. As shown in the table of Fig. \ref{fig:vec_random}, classification has a positive Pearson Correlation coefficient, but regression has zero or small negative correlations. Our conclusion is further substantiated. We suppose that the reason behind the performance drop of vec-FactorVAE is: if the regularization strength is too large for FactorVAE, there is a significant performance drop. This also can be observed in Fig. 5 from ~\cite{FactorVAE}. We also provide the experiments when $\gamma \leq 5$ in Appendix D.

\textbf{Implications} For a certain type of vector-based model, the compositional generalization (classification) ability is positive correlate to the disentanglement. However, the disentanglement has only limited influence on regression metrics. We suppose this is because the regression is conducted on factors with large amounts of values (e.g., azimuth with 15 values on Shapes3D), which contains more information and will be more sensitive to vector size.
The regression is only positively correlated with the vector size of the representation. 

\subsection{Does Increase the Representation Dimension Affect the Metrics?}

It is worth questioning whether directly increasing the feature dimensions affects the metrics or whether the model learns a good representation. To answer this question, we design the following experiments that evaluate different artificial representations to exclude the possibility that the gain of the vector-based model comes from the increase of the feature dimensions directly. We first evaluate the representation derived from the ground truth values, which is ideally a perfect representation of disentanglement and generalization. We name such representation as ideal representation.

The first question that arises is whether scalar-based representation can achieve high performance on two kinds of metrics. Since the ground truth values are given by scalars, one can construct the ideal representation in the following: $(i)$ We construct the ideal scalar-based representation by directly normalizing the ground truth values of corresponding factors. $(ii)$ For the ideal vector-based representation, we first normalize the ground truth values of the factors. We then multiply the normalized values with a random embedding to directly increase the vector size of the corresponding factor. Note that the embedding is shared across the training and testing set. As shown in Tab. \ref{tbl:ideal_exp} (a), no matter whether the representation is vector-based or scalar-based, the representation performs perfectly both on generalization and disentanglement.
\begin{table}[t]
\caption{Directly increase the representation evaluation experiment. We evaluate the following ideal and learned representation in terms of compositional generalization and disentanglement.}
\begin{center}
\begin{tabular}{cc}
\resizebox{0.4\linewidth}{!}{

\begin{tabular}{lc@{\hspace{3em}}c@{\hspace{3em}}c}
\toprule
\multicolumn{1}{c}{Method} & R2 &ACC &  DCI \\
\midrule
\multicolumn{4}{c}{\textit{scalar-based:}} \\
\midrule
Ideal             & $1.00$ & $0.99$ & $0.99$ \\
Shifted     & $0.46$ & $0.75$& $0.99$  \\
Matrix      & $0.99$ & $1.00$ & $0.18$ \\
Matrix $+$ shifted     & $0.00$ & $0.45$ & $0.20$ \\
\midrule
\multicolumn{4}{c}{\textit{vector-based:}} \\
\midrule
Ideal             & $1.00$ & $0.99$ & $0.99$  \\
Shifted     & $0.38$ & $0.50$& $0.99$  \\
Matrix      & $0.99$ & $1.00$ & $0.17$ \\

Matrix $+$ shifted     & $0.00$ & $0.45$ & $0.17$ \\

\bottomrule
\end{tabular} } &
\resizebox{0.45\linewidth}{!}{
\begin{tabular}{lc@{\hspace{3em}}c@{\hspace{3em}}c}
\toprule
\multicolumn{1}{c}{Method} & $R^2$ &ACC &  DCI \\
\midrule
\multicolumn{4}{c}{\textit{scalar-based:}} \\
\midrule
$\beta$-TCVAE     & $0.30$ & $0.56$ & $0.61$ \\
FactorVAE     & $0.49$ & $0.48$ & $0.50$  \\
\midrule
\multicolumn{4}{c}{\textit{vector-based:}} \\
\midrule
$\beta$-TCVAE embed     & $0.30$ & $0.56$& $0.61$  \\
$\beta$-TCVAE repeat     & $0.30$ & $0.56$& $0.61$  \\

FactorVAE embed     & $0.49$ & $0.48$& $0.50$ \\
FactorVAE repeat     & $0.49$ & $0.48$& $0.50$  \\

\bottomrule
\end{tabular} }\\
(a) Ideal representation & (b) Learned representation
\end{tabular}
\label{tbl:ideal_exp}
\end{center}
\vspace{-1.5em}

\end{table}

We use the following two ways to corrupt the disentanglement and generalization abilities of ideal representations:  
$(i)$ To corrupt the generalization ability, we apply two different linear operations ($y = \alpha x + \beta$, where $\alpha, \beta$ are scalars) to the ideal representation on the training and testing set. $(ii)$ To corrupt the disentanglement ability, we multiply the ideal representation with a random invertible matrix to entangle the representation of different factors. In Tab. \ref{tbl:ideal_exp} (a), we dose not observe that the performance of vector-based representation is significantly improved.

Although the corruption above also sheds light on the question above, there is a gap between the learned representation and artificially corrupted representation. To study the question in a non-ideal setting, we propose two ways that directly map the scalar-based representation to a vector-based one. Similar to the ideal vector-based representation: $(i)$ We multiply the learned scalar-based representation of VAE with a random embedding of the corresponding factor. $(ii)$ We directly repeat the scalar of the learned representation of VAE. As shown in Tab. \ref{tbl:ideal_exp} (b), we use $\beta$-TCVAE and FactorVAE as examples. The metrics of vector-based representation do not significantly improve compared to the original scalar-based one.

\textbf{Implications} The reason behind the performance gain is not that directly increasing the representation dimension leads to performance gains. We also verified that if the representation is similar to the ideal scalar representation, we also can obtain the perfect generalization performance. 



\section{Limitation of Our Study}

Our study is built upon prior work~\cite{xu2022compositional}, in which experiments are conducted on the two proposed metrics. However, to the best of our knowledge, these metrics are the unique ones directly evaluating compositional generalization with random train-test splitting. Although our experiments are conducted on the synthetic (Shapes3D) and realistic (MPI3D) datasets, the factors within these datasets are relatively simple. Moreover, the number of factors is known and relatively small. While we demonstrate a set of models with strong disentanglement and generalization performance, the potential applications of these models still need to be explored.

\section{Conclusions and Discussions}

In this paper, we proposed unifying existing models into vector-based and scalar-based disentanglement methods. Specifically, we modify scalar-based disentanglement works ($\beta$-TCVAE, FactorVAE, and SAE) to be vector-based. Additionally, we also modify the vector-based method (VCT) to be scalar-based. We study these disentangled representation learning works in terms of disentanglement and compositional generalization abilities. Based on this study, we present an interesting finding: vector-based representation (using a vector instead of a scalar to represent a concept) is the key to empowering good disentanglement and strong compositional generalization. This finding first highlights the importance of vector-based disentangled representation. 
We observe that increasing the number of dimensions of vector-based representation improves the compositional generalization. Focusing on vector-based disentanglement, we reconsider the relationship between disentanglement and compositional generalization. We find that classification is surprisingly positively correlated to disentanglement for vector-based methods.
We hope that our study encourages further research on compositional generalization and vector-based disentanglement methods. Interesting future directions include working on more complex large-natural datasets with more factors and discussing the relationship between disentanglement on real-world, large-scale natural datasets and compositional generalization. The potential negative societal impacts are malicious uses.

{
\small
\bibliographystyle{plainnat}
\bibliography{neurips_2023}
}


\end{document}


\maketitle

\section{More Details of Loss Function}
\subsection{KL Divergence for Vector-based VAE Methods}
\begin{thm} Let $x$ be an $n$-dimensional random vector. Assume $x$ is sampled from either one of the two multivariate normal distributions $p$ or $q$, which are specified by mean vector $\mu_p, \mu_q\in \mathbb{R}^n$, and variance matrix $\Sigma_p, \Sigma_q \in \mathbb{R}^{n\times n}$, respectively, as
$x\sim p(\mu_p,\Sigma_p)$ and $x\sim p(\mu_q,\Sigma_q)$. Then, the Kullback-Leibler (KL) divergence of $p$ from $q$ is given by
\begin{equation}
    \mathbf{KL}(p(x)||q(x)) = \frac{1}{2}\left[(\mu_q-\mu_p)^T\Sigma_q^{-1}(\mu_q- \mu_p) + tr(\Sigma_q^{-1}\Sigma_p) - \log\frac{|\Sigma_p|}{|\Sigma_q| } - n\right].
\end{equation}
\end{thm}
As introduced in Section 4.1 of the main paper, we use a spherical Gaussian for each latent representation unit in the vectorized VAE-based model. For the $i$-th representation unit of VAE, we use a mean vector $\mu_i$ and a scalar $\sigma_i$ to characterize the spherical Gaussian $q(z_i|x) = \mathcal{N}(\mu_i,\sigma_i I)$. Therefore, according to the theorem above, the KL divergence between the posterior and prior is as follows:
\begin{equation}
\begin{array}{ll}
     \mathbf{KL}(q(z|x)||p(z))& = \sum_i\mathbb{E}_{q(z_i|x)}(\log q(z_i|x) - \log p(z_i)) \\
    &= \sum_i0.5\left(\sum_j\mu_{ij}^2 + D\sigma_i - D\log \sigma_i - D\right) \\
    & = 0.5\left(1/D \sum_i\sum_j\mu_{ij}^2 +\sum_i\sigma_i - \sum_i\log \sigma_i - m\right)D \\
     & = 0.5\left(1/D \sum_{ij}\mu_{ij}^2 +\sum_i(\sigma_i - \log \sigma_i) - m\right)D
\end{array} 
\end{equation}
where $m$ is the number of latent representation units of the VAE-based model. We use a spherical Gaussian to characterize each unit. And $\mu_{ij}$ indicates the $j$-th entry of the mean vector $\mu_i$.
\subsection{The total correlation for vec-\texorpdfstring{$\beta$}--TCVAE}
Recall that $\beta$-TCVAE utilizes a sampling method to estimate the total correlation. Specifically, in order to estimate the expectation of $\log q(z)$, we first sample a mini-batch of images: $\{x^1, x^2,\dots,x^M\}$, and estimate $\mathbb{E}_{q(z)}[q(z)]$ as follows:
\begin{equation}
    \mathbb{E}_{q(z)}[q(z)] \approx \frac{1}{M}\sum_{k=1}^M\left[\log \sum_{l=1}^M\exp (E(z(x_l)|x_k)) - \log{MK})\right],
\end{equation}
where $E(z(x^l)|x^k) = \log q(z(x^l)|x^k)$ is the log density of the distribution $q(z(x^l)|x^k)$, and $K$ is the number of samples of the dataset. We use $z(x^l)$ to indicate the representation derived by sample $x^l$. The total correlation can be estimated by the following equation:
\begin{equation}
    \mathbf{KL}\left(q(z)||\Pi_i q(z_i)\right) = \mathbb{E}_{q(z)}\left[\log q(z) - \sum_i\log q(z_i)\right].
\end{equation}
where $z_i$ denotes the $i$-th latent representation unit.

For the vector-based case, we know that it is important to calculate the log density of the distribution to estimate the total correlation. The log density of spherical Gaussian can be computed as follows:
\begin{equation}
    E(z_i|x^k) = \sum_{j=1}^DE(z_{ij}|x^k),
    \label{eq:sum}
\end{equation}
which indicates that $E(z_i|x^k)$ is a large negative scalar, since $E(z_{ij}|x^k) < 0$, the value of the probability $q(z_i|x^k) = \exp(E(z_i|x^k))$ is expected to be very close to $0$, which requires a high degree of precision. However, the limited precision of  GPUs produces relatively large truncation errors when dealing with these small numerical values. Therefore, the original estimator is no longer fit for vector-based $\beta$-TCVAE, and the total correlation is intractable in this way. Therefore, we use the following way to substitute the estimated total correlation instead.

We constrain a necessary condition of the independence of the marginal distribution $q(z_i)$. Specifically, the total correlation is computed by the same entry of different representation units, i.e., $z_{ij}$ and $z_{kj}$ are independent, for $i\neq k$. We then can derive a new equation for the total correlation:
\begin{equation}
     \mathcal{L}_{\beta\textrm{-TCVAE}}
      = \sum_j(\mathbf{KL}\left(q(z_j)||\Pi_i q(z_{ij})\right)),
      \label{eq:new_tc}
\end{equation}
where $\mathcal{L}_{\beta\textrm{-TCVAE}}$ is the regularization term of the vec-$\beta$-TCVAE loss function, which plays the role of total correlation $\mathbf{KL}(q(z)||\Pi_i q(z_{i}))$. Similarly, we also normalize it by dividing the vector dimension $D$, and then the value of the total correlation regularization is comparable to the original one of $\beta$-TCVAE.

From Eq.\ref{eq:new_tc}, we see that each term in the summation is the total correlation of one-dimensional Gaussians and there is no need for compute Eq.\ref{eq:sum} for each unit. Therefore, the abovementioned issue does not exist. Since the regularization constraint of vec-$\beta$-TCVAE is only a necessary condition for the original model, the disentanglement performance of  vec-$\beta$-TCVAE is relatively poor. In the following, we provide the theorem and proof of the necessity. For simplicity, we use two random vectors as an example.

\begin{thm} If random vectors $\mathbf{z}_1= \{z_{11},\dots, z_{1D}\}$ and $\mathbf{z}_2= \{z_{21},\dots, z_{2D}\}$ are independent, the corresponding random variables $z_{1i}$ and $z_{2i}$ are independent, where $1\leq i\leq D$.
\end{thm}

\textit{Proof}: Given that random vectors $\mathbf{z}_1$ and $\mathbf{z}_2$ are independent, their joint probability distribution can be expressed as the product of their marginal probability distributions:

\begin{equation}
p(\mathbf{z}_1, \mathbf{z}_2) = p(\mathbf{z}_1)p(\mathbf{z}_2).
\end{equation}

Now let's consider any two random variables: $z_{1i}$ from $\mathbf{z}_1$ and $z_{2i}$ from $\mathbf{z}_2$. We want to show that $p(z_{1i}, z_{2i}) = p(z_{1i})p(z_{2i})$. First, we can write the joint probability distribution of $z_{1i}$ and $z_{2i}$ using the joint distribution of the vectors $\mathbf{z}_1$ and $\mathbf{z}_2$:
\begin{equation}
p(z_{1i}, z_{2i}) = \int_{\forall z_{1k} \ne z_{1i}, \forall z_{2k} \ne z_{2i}} p(\mathbf{z}_1, \mathbf{z}_2)dz_{1k}dz_{2k}.
\end{equation}

Since $\mathbf{z}_1$ and $\mathbf{z}_2$ are independent, we can substitute their product of marginals:
\begin{equation}
p(z_{1i}, z_{2i}) = \int_{\forall z_{1k}\ne z_{1i}}\int_{\forall z_{2k}\ne z_{2i}}p(\mathbf{z}_1) p(\mathbf{z}_2)dz_{1k}dz_{2k}.
\end{equation}

Notice that the summation is not affected by the values of $z_{1i}$ and $z_{2i}$. Therefore, we can separate the summation into two parts, one for each vector:
\begin{equation}
    p(z_{1i}, z_{2i}) = \left(\int_{\forall z_{1k}\ne z_{1i}}p(\mathbf{z}_1)dz_{1k}\right) \left(\int_{\forall z_{2k}\ne z_{2i}}p(\mathbf{z}_2)dz_{2k}\right).
\end{equation}

By definition, the summation of all the marginal probabilities for each vector is equal to the marginal probability of $z_{1i}$ and $z_{2i}$:

\begin{equation}
    p(z_{1i}, z_{2i}) = p(z_{1i})p(z_{2i}).
\end{equation}

Thus, if random vectors $\mathbf{z}_1$ and $\mathbf{z}_2$ are independent, then the corresponding random variables $z_{1i}$ and $z_{2i}$ are also independent.

\section{More Implementation Details}
\textbf{Model architectures}. In our experiments, we follow \cite{DisCo}\footnote{https://github.com/xrenaa/DisCo} and use the convolutional auto-encoder architectures for VAE-based methods, which is presented in Table \ref{tbl:encode_type} and \ref{tbl:decode_type}. 
For vec-VCT\footnote{https://github.com/ThomasMrY/VCT}, we use $m$ MLPs to reduce the dimension of $m$ concept tokens from 256 to $D$, which presented in Table \ref{tbl:MLPs} (a).  For vec-VCT*, we do not use these MLPs to maintain the dimension as $D = 256$. We follow ~\cite{leebstructure} to use the same architecture of SAE\footnote{code is in https://openreview.net/forum?id=ue4CArRAsct}, and change the MLP in the encoder and Str-Tfm layer to produce the vector-based representation, as shown in Table \ref{tbl:MLPs} (b) and (c). 

\begin{table}[ht]
\begin{center}
\caption{Encoder $\mathcal{E}$ used in vec-VAE-based methods. We use $m$ to denote the number of latent representation units, and $D$ to denote the vector dimension. We set $D=1$ for scalar based methods.}
\begin{tabular}{l}
\toprule
Conv $7 \times 7 \times 3 \times 64$, stride $= 1$ \\
ReLu \\
Conv $4 \times 4 \times 64 \times 128$, stride $= 2$ \\
ReLu \\
Conv $4 \times 4 \times 128 \times 256$, stride $= 2$ \\
ReLu \\
Conv $4 \times 4 \times 256 \times 256$, stride $= 2$ \\
ReLu \\
Conv $4 \times 4 \times 256 \times 256$, stride $= 2$ \\
ReLu \\
FC $4096 \times 256$ \\
ReLu \\
FC $256 \times 256$ \\
ReLu \\
FC $256 \times 2mD$ \\
\bottomrule
\end{tabular}
\label{tbl:encode_type}
\end{center}
\end{table}

\begin{table}[ht]
\caption{Decoder $\mathcal{D}$ architecture used in vec-VAE-based methods.}
\begin{center}
\begin{tabular}{l}
\toprule
FC $mD \times 256$ \\
ReLu \\
FC $256 \times 256$ \\
ReLu \\
FC $256 \times 4096$ \\
\midrule
ConvTranspose $4 \times 4 \times 256 \times256$, stride $= 2$ \\
ReLu \\
ConvTranspose $4 \times 4 \times 256 \times256$, stride $= 2$ \\
ReLu \\
ConvTranspose $4 \times 4 \times 256 \times128$, stride $= 2$ \\
ReLu \\
ConvTranspose $4 \times 4 \times 128 \times64$, stride $= 2$ \\
ReLu \\
ConvTranspose $7 \times 7 \times 64\times 3$, stride $= 2$ \\
\bottomrule
\end{tabular}
\end{center}
\label{tbl:decode_type}

\end{table}

\begin{table}[ht]
\caption{MLP in vector-based models. (a) We use $m$ MLPs in vec-VCT. (b) We change the output of SAE's MLP from $m$ to $mD$. (c) We change the input of SAE's Str-Tfm layer from $1$ to $D$.}
\begin{tabular}
{m{5cm}<{\centering}m{4cm}<{\centering}m{3cm}}
\begin{center}
\begin{tabular}{l|l}
\toprule
Encoder MLP & Decoder MLP \\
\midrule
FC $256 \times 128$ & FC $D \times 128$\\
ReLu & ReLu\\
FC $128 \times D$ & FC $128 \times 256$\\
\bottomrule
\end{tabular}
\end{center} & \begin{center}
\begin{tabular}{l}
\toprule
FC $256 \times 256$ \\
ReLu \\
FC $256 \times 256$ \\
ReLu \\
FC $256 \times mD$ \\
\bottomrule
\end{tabular}
\end{center} & \begin{center}
\begin{tabular}{l}
\toprule
FC $D \times 64$ \\
ReLu\\
FC $64 \times 128$ \\
ReLu\\
FC $128 \times (64\times2)$ \\
\bottomrule
\end{tabular}
\end{center} \\
(a) MLP in vec-VCT & (b) MLP in SAE encoder & (c) MLP in Str-Tfm  
\end{tabular}
\vspace{0.5em}
\label{tbl:MLPs}
\end{table}

\textbf{Evaluation metrics}. For evaluating the compositional generalization, we follow~\cite{xu2022compositional} to use the implementations\footnote{https://github.com/wildphoton/Compositional-Generalization} in scikit-learn (version 0.22) for the linear models of compositional generalization metrics. Specifically, we use the function \texttt{linear\_model.LogisticRegressionCV} with default settings for classification, and use function \texttt{linear\_model.RidgeCV} with alphas=$\{0, 0.01, 0.1, 1.0, 10\}$ for regression. There are $N = 500$ samples for training models of the metrics. We use the same metrics configurations as in ~\cite{locatello2019challenging} to evaluate the disentanglement performance. Since the representation used is vector-based and can not be evaluated directly, we thus follow ~\cite{yang2022visual, du2021unsupervised} to perform PCA as post-processing on the representation and evaluate the performance with these metrics.

\textbf{Computational cost}. No matter vector-based or scalar-based methods, we use an Nvidia V100 16G as the compute resource for conducting the experiments.

\textbf{Reproducibility}. Our code will be released upon acceptance.

\section{More Results on Metrics}
The main paper only presents the important experiment results to support the key findings. We provide more results with disentanglement metrics $\beta$-VAE score and MIG.
In order to analyze the influence of the training-testing splitting ratio, we provide more results with different ratios.

\begin{table*}[t]
\vspace{-1em}
\caption{Comparisons of more disentanglement metrics between the scalar-based and vector-based methods (mean $\pm$ std, higher is better).  For vec-VCT*, $D=256$ is the same as ~\citet{yang2022visual}. }
\begin{center}
\resizebox{0.6\textwidth}{!}{
\begin{tabular}{ccc|cc}
\toprule
\multirow{2}*{\textbf{Method}} & \multicolumn{2}{c}{Shapes3D} & \multicolumn{2}{c}{MPI3D} \\
\cmidrule(lr){2-5}
& $\beta$-VAE & MIG & $\beta$-VAE & MIG  \\
\midrule
\multicolumn{5}{c}{\textit{Scalar-based:}} \\
\midrule
FactorVAE & $0.89 \pm 0.05$ & $0.21 \pm 0.12$ & $0.51 \pm 0.05$ & $0.11 \pm  0.05$ \\
$\beta$-TCVAE & $0.88 \pm 0.07$ & $0.37 \pm 0.19$ & $0.46 \pm 0.03$ & $0.11 \pm  0.05$ \\
SAE & $0.99 \pm 0.01$ & $0.32 \pm 0.10$ & $0.79 \pm 0.03$ & $0.18 \pm  0.06$ \\
VCT & $0.97 \pm 0.05$ & $0.40 \pm 0.11$ & $0.78 \pm 0.05$ & $0.26 \pm  0.05$ \\
\midrule
\multicolumn{5}{c}{\textit{Vector-based:}} \\
\midrule
vec-FactorVAE & $0.98 \pm 0.02$ & $0.25 \pm 0.07$ & $0.46 \pm 0.07$ & $0.10 \pm  0.05$ \\
vec-$\beta$-TCVAE & $0.94 \pm 0.04$ & $0.12 \pm 0.07$ & $0.49 \pm 0.04$ & $0.04 \pm  0.01$ \\
vec-SAE & $0.96 \pm 0.05$ & $0.22 \pm 0.09$ & $0.70 \pm 0.06$ & $0.15 \pm  0.09$ \\
vec-VCT & $0.99 \pm 0.02$ & $0.42 \pm 0.10$ & $0.77 \pm 0.05$ & $0.35 \pm  0.07$ \\
vec-VCT* & $1.00 \pm 0.00$ & $0.44 \pm 0.08$ & $0.74 \pm 0.05$ & $0.33 \pm  0.06$ \\

\bottomrule
\end{tabular}}
\end{center}
\vspace{-1.5em}
\label{tbl:dis_qnti}
\end{table*}
\textbf{Results of more metrics}.
Table \ref{tbl:dis_qnti} presents the results of additional disentanglement metrics in this section: $\beta$-VAE score, MIG. Consistently to our results in the main paper, we observe that the vec-models have comparable or even better performance on some metrics. The results further support that the vec-model can simultaneously process the compositional generalization and disentanglement.

\begin{figure}[t]
\centering
\begin{tabular}{c@{\hspace{0.1em}}c@{\hspace{0.1em}}c}
\multicolumn{3}{c}{\includegraphics[width=0.54\linewidth]{appendix/shapes3d/legend_ratio.png}}\\
\includegraphics[width=0.32\linewidth]{appendix/shapes3d/R2_vec_ratio.png} & \includegraphics[width=0.32\linewidth]{appendix/shapes3d/ACC_vec_ratio.png} & 
\includegraphics[width=0.32\linewidth]{appendix/shapes3d/DCI_vec_ratio.png}  \\
\includegraphics[width=0.32\linewidth]{appendix/shapes3d/Factor_vec_ratio.png} &
\includegraphics[width=0.32\linewidth]{appendix/shapes3d/MIG_vec_ratio.png}&
\includegraphics[width=0.32\linewidth]{appendix/shapes3d/Beta_vec_ratio.png}
\end{tabular}
\caption{Generalization and disentanglement performance vs training-testing splitting ratio on Shapes3D. Three vector-based methods (with vector size $D=64$, regualarization strength $\gamma=10.0$) are evaluated. vec-VCT, vec-betaTCVAE, and vec-FactorVAE use varying training-testing splitting ratios $\{3:7,1:9,5:95\}$.}
\label{fig:vec_split}
\vspace{-1em}
\end{figure}

\begin{figure}[t]
\centering
\begin{tabular}{c@{\hspace{0.1em}}c@{\hspace{0.1em}}c}
\multicolumn{3}{c}{\includegraphics[width=0.54\linewidth]{appendix/mpi3d/legend_lam.png}}\\
\includegraphics[width=0.32\linewidth]{appendix/mpi3d/R2_vec_lam.png} & \includegraphics[width=0.32\linewidth]{appendix/mpi3d/ACC_vec_lam.png} & 
\includegraphics[width=0.32\linewidth]{appendix/mpi3d/DCI_vec_lam.png}  \\
\includegraphics[width=0.32\linewidth]{appendix/mpi3d/Factor_vec_lam.png} &
\includegraphics[width=0.32\linewidth]{appendix/mpi3d/MIG_vec_lam.png}&
\includegraphics[width=0.32\linewidth]{appendix/mpi3d/Beta_vec_lam.png}
\end{tabular}
\caption{Generalization and disentanglement performance vs regularization strength on MPI3D. Two vector-based methods (with vector size $D=64$) are evaluated. vec-betaTCVAE and vec-FactorVAE use varying regularization strength $\gamma \in \{5,10,20\}$.}
\label{fig:vec_lam_mpi}
\vspace{-1em}
\end{figure}
\textbf{Results of different training-testing splitting ratio} In the main paper, we follow \cite{xu2022compositional} to conduct the experiments mainly with a training-testing splitting ratio (split ratio) of $1:9$. In this section, we also provide the results of the experiments with three different ratios: $3:7$, $1:9$, and $5:95$. As shown in Figure \ref{fig:vec_split}, the split ratio has limited influence on the compositional metrics and disentanglement metrics, which is also observed in ~\cite{xu2022compositional}. However, their experiments are conducted on scalar-based methods.

\section{More Results on Regularization Strength}
In this section, we provide results that regularization strength is less than five as the supplement results. We also provide the results on the MPI3D dataset to present more support for our findings.
\begin{figure}[t]
\centering
\begin{tabular}{c@{\hspace{0.1em}}c@{\hspace{0.1em}}c}
\multicolumn{3}{c}{\includegraphics[width=0.54\linewidth]{appendix/shapes3d/legend_lam.png}}\\
\includegraphics[width=0.32\linewidth]{appendix/shapes3d/R2_vec_lam.png} & \includegraphics[width=0.32\linewidth]{appendix/shapes3d/ACC_vec_lam.png} & 
\includegraphics[width=0.32\linewidth]{appendix/shapes3d/DCI_vec_lam.png}  \\
\includegraphics[width=0.32\linewidth]{appendix/shapes3d/Factor_vec_lam.png} &
\includegraphics[width=0.32\linewidth]{appendix/shapes3d/MIG_vec_lam.png}&
\includegraphics[width=0.32\linewidth]{appendix/shapes3d/Beta_vec_lam.png}
\end{tabular}
\caption{Generalization and disentanglement performance vs regularization strength on Shapes3D. Two vector-based methods (with vector size $D=64$) are evaluated. vec-betaTCVAE and vec-FactorVAE use varying regularization strength $\gamma \in \{0.1,1.0,2.0,4.0\}$.}
\label{fig:vec_lam_shapes3d}
\vspace{-1em}
\end{figure}

\textbf{FactorVAE when $\gamma \leq 5$}. Figure \ref{fig:vec_lam_shapes3d} shows the results of vec-FactorVAE and vec-$\beta$-TCVAE when the regularization strength $\gamma \in \{0.5,1,2,4\}$. We can observe that the results support the explanation in the main paper: if the regularization strength is small for vec-FactorVAE, there also is a significant improvement in the compositional generalization and disentanglement performance as the increase of $\gamma$. We also observe that the phenomenon is consistent between vec-FactorVAE and vec-$\beta$-TCVAE.

\textbf{Results on MPI3D}. To further support our conclusion on the regularization strength $\gamma$ of vector-based methods, we conduct the same experiment on MPI3D. As shown in Figure \ref{fig:vec_size_mpi}, the results 
 are consistent with our results in the main paper. We observe that as the regularization strength
increase, both the compositional generalization and the disentanglement drops for vec-FactorVAE but improves for vec-$\beta$-TCVAE. The additional metrics (MIG and $\beta$-VAE score) are consistent with other metrics.

\textbf{Pearson correlation coefficient on MPI3D}. In the main paper, the experiments reveal the relation between compositional generalization and disentanglement based on the results of Shapes3D. We also provide such evidence on MPI3D here. From the results in Figure \ref{fig:vec_random}, we observe consistent patterns: the classification performance is positively correlated to the disentanglement, but there is no significant correlation between the regression and disentanglement performance.

\begin{figure}[t]
\centering
\begin{tabular}
{m{3cm}<{\centering}m{3cm}<{\centering}m{3cm}<{\centering}m{3.5cm}}
\includegraphics[width=\linewidth]{appendix/mpi3d/bTC_vec_random_R2.png} & \includegraphics[width=\linewidth]{appendix/mpi3d/factor_vec_random_R2.png} & 
\includegraphics[width=\linewidth]{appendix/mpi3d/VCT_vec_random_R2.png} & \multirow{2}*{\resizebox{0.95\linewidth}{!}{ 
\begin{tabular}{lc}
\toprule
\multicolumn{1}{c}{Method} & Correlation\\
\midrule
\multicolumn{2}{c}{\textit{Regression:}} \\
\midrule
vec-$\beta$-TCVAE             & $0.049$   \\
vec-FactorVAE               & $-0.196$ \\ 
vec-VCT     & $0.159$  \\
\midrule
\multicolumn{2}{c}{\textit{Classification:}} \\
\midrule
vec-$\beta$-TCVAE             & $0.471$   \\
vec-FactorVAE               & $0.481$ \\ 
vec-VCT    & $0.768$  \\
\bottomrule
\end{tabular}}}\\
\includegraphics[width=\linewidth]{appendix/mpi3d/bTC_vec_random_ACC.png} & \includegraphics[width=\linewidth]{appendix/mpi3d/factor_vec_random_ACC.png} & 
\includegraphics[width=\linewidth]{appendix/mpi3d/VCT_vec_random_ACC.png} \\
vec-$\beta$-TCVAE & vec-FactorVAE  & vec-VCT & \quad Pearson Correlation \\
\end{tabular}
\caption{Generalization performance vs disentanglement performance on the MPI3D with $D=64$ and $\gamma=10$. The Pearson correlation coefficient is calculated in the table. Orange data points represent instances of the same vector-based method trained using different random seeds.}
\label{fig:vec_random}
\end{figure}
\begin{figure}[t]
\centering
\begin{tabular}{c@{\hspace{0.1em}}c@{\hspace{0.1em}}c}
\multicolumn{3}{c}{\includegraphics[width=0.54\linewidth]{appendix/shapes3d/legend_size.png}}\\
\includegraphics[width=0.32\linewidth]{appendix/shapes3d/R2_vec_size.png} & \includegraphics[width=0.32\linewidth]{appendix/shapes3d/ACC_vec_size.png} & 
\includegraphics[width=0.32\linewidth]{appendix/shapes3d/DCI_vec_size.png}  \\
\includegraphics[width=0.32\linewidth]{appendix/shapes3d/Factor_vec_size.png} &
\includegraphics[width=0.32\linewidth]{appendix/shapes3d/MIG_vec_size.png}&
\includegraphics[width=0.32\linewidth]{appendix/shapes3d/Beta_vec_size.png}
\end{tabular}
\caption{Generalization and disentanglement performance vs regularization strength on Shapes3D. Two vector-based methods (with regularization strength $\gamma=10$) are evaluated. vec-betaTCVAE and vec-FactorVAE use varying vector size $D \in \{4,8,16,24\}$.}
\label{fig:vec_size}
\vspace{-1em}
\end{figure}

\begin{figure}[t]
\centering
\begin{tabular}{c@{\hspace{0.1em}}c@{\hspace{0.1em}}c}
\multicolumn{3}{c}{\includegraphics[width=0.54\linewidth]{appendix/mpi3d/legend_size.png}}\\
\includegraphics[width=0.32\linewidth]{appendix/mpi3d/R2_vec_size.png} & \includegraphics[width=0.32\linewidth]{appendix/mpi3d/ACC_vec_size.png} & 
\includegraphics[width=0.32\linewidth]{appendix/mpi3d/DCI_vec_size.png}  \\
\includegraphics[width=0.32\linewidth]{appendix/mpi3d/Factor_vec_size.png} &
\includegraphics[width=0.32\linewidth]{appendix/mpi3d/MIG_vec_size.png}&
\includegraphics[width=0.32\linewidth]{appendix/mpi3d/Beta_vec_size.png}
\end{tabular}
\caption{Generalization and disentanglement performance vs regularization strength on MPI3D. Two vector-based methods (with regularization strength $\gamma=10$) are evaluated. vec-betaTCVAE and vec-FactorVAE use varying vector size $D \in \{1,2,32,64\}$.}
\label{fig:vec_size_mpi}
\vspace{-1em}
\end{figure}

\section{More Results on Vector Size}
In this section, we provide results that vector size between 2 and 32 as the supplement results for the main paper. We also provide the results on the MPI3D dataset.

\textbf{More results on Shapes3D}. In the main text, we only considered some specific and important vector dimension values, which are $\{1,2,32,64\}$. However, due to the large span between $2$ and $32$, one may be interested in other values within the interval. Therefore, we also experimented with the values within the interval. Figure \ref{fig:vec_size} shows the results when vector size is $\{4, 8, 16, 24\}$. We also observe that as the increase vector dimension (size) increased, both the compositional and the disentanglement consistently improved, which provides further evidence for the conclusion in Section 6.2.

\textbf{Results on MPI3D}. In order to further support our conclusion on vector size of vector-based methods, we conduct the same experiment on MPI3D. As shown in Figure \ref{fig:vec_size_mpi}, we see that as the vector dimension (size) increases, the compositional generalization performance of the models improves consistently. The disentanglement performance of certain models (e.g., VCT) also shows improvement. However, some models do not significantly exhibit such improvements in some of the metrics. The reason may be that these methods have poor performance on these metrics on MPI3D. Consequently, changing the vector dimension does not result in substantial performance changes for these models.

{
\small
\bibliographystyle{plainnat}
\bibliography{neurips_2023}
}